\documentclass[10pt,journal,compsoc]{IEEEtran}

\ifCLASSOPTIONcompsoc
  \usepackage[nocompress]{cite}
\else
  \usepackage{cite}
\fi

\ifCLASSINFOpdf
  \usepackage[pdftex]{graphicx}

\usepackage{amsmath}

\usepackage{url}

\usepackage{amssymb}
\usepackage{color}

\begin{document}

\title{A Deeper Look into DeepCap}

\author{Marc~Habermann,~\IEEEmembership{MPII,}
        Weipeng~Xu,~\IEEEmembership{MPII,}
        Michael~Zollhoefer,~\IEEEmembership{Stanford University,}
        Gerard~Pons-Moll,~\IEEEmembership{MPII,}
        and~Christian~Theobalt,~\IEEEmembership{MPII}
\thanks{Manuscript revised June, 2021.}}

\markboth{TO APPEAR IN IEEE TRANSACTIONS ON PATTERN ANALYSIS AND MACHINE INTELLIGENCE, 2021}%
{Habermann \MakeLowercase{\textit{et al.}}: A Deeper Look into DeepCap}

\IEEEspecialpapernotice{(Invited Paper)}

\IEEEtitleabstractindextext{%
\begin{abstract}
\label{sec:abstract}
Human performance capture is a highly important computer vision problem with many applications in movie production and virtual/augmented reality.
Many previous performance capture approaches either required expensive multi-view setups or did not recover dense space-time coherent geometry with frame-to-frame correspondences.
We propose a novel deep learning approach for monocular dense human performance capture.
Our method is trained in a weakly supervised manner based on multi-view supervision completely removing the need for training data with 3D ground truth annotations.
The network architecture is based on two separate networks that disentangle the task into a pose estimation and a non-rigid surface deformation step.
Extensive qualitative and quantitative evaluations show that our approach outperforms the state of the art in terms of quality and robustness.
This work is an extended version of \cite{habermann20} where we provide more detailed explanations, comparisons and results as well as applications.
\end{abstract}

}

\maketitle
\IEEEdisplaynontitleabstractindextext
\IEEEpeerreviewmaketitle

\IEEEraisesectionheading{\section{Introduction}\label{sec:intro}}
%
%
%
\par
\IEEEPARstart{C}{apturing} the space-time coherent geometry of entire humans including also their everyday clothing, which is also known as human performance capture, is extensively used nowadays for movie production and game development.
With these tools, real humans can be ported into a virtual world and fused into the process of creating virtual content.
But also other applications such as virtual try-on, telepresence and virtual and augmented reality applications can strongly benefit from dense capture methods.
Especially the latter mentioned applications are mainly used by non-expert users, who do not own dense multi-capture setups.
Thus, to democratize these technologies the hardware setup has be as simple and accessible as possible, which means ideally a \textit{single color camera} is sufficient to capture the entire deforming surface of a human.
While there is extensive research regarding monocular skeleton-only tracking, only few works have targeted the capture of the dense deforming geometry of the entire human from a single color stream.
Nonetheless, densely capturing entire humans from a single view is still far from being solved while it is indispensable for creating higher fidelity and photo-real characters.
%
%
%
\par
Some works~\cite{bray06,brox06,de08, vlasic08, gall09, vlasic09, brox10,cagniart10,liu11,wu13, mustafa15,pons15, pons17} have focused on capturing entire humans from multi-camera setups that typically also include a green screen.
While they achieve a superior quality, the hardware requirement makes it almost impossible to use them in other locations like an outdoor film set and restricts the usage to companies that can afford such an expensive setup.
%
%
\begin{figure}[t]
	\begin{center}
		\includegraphics[width=\linewidth]{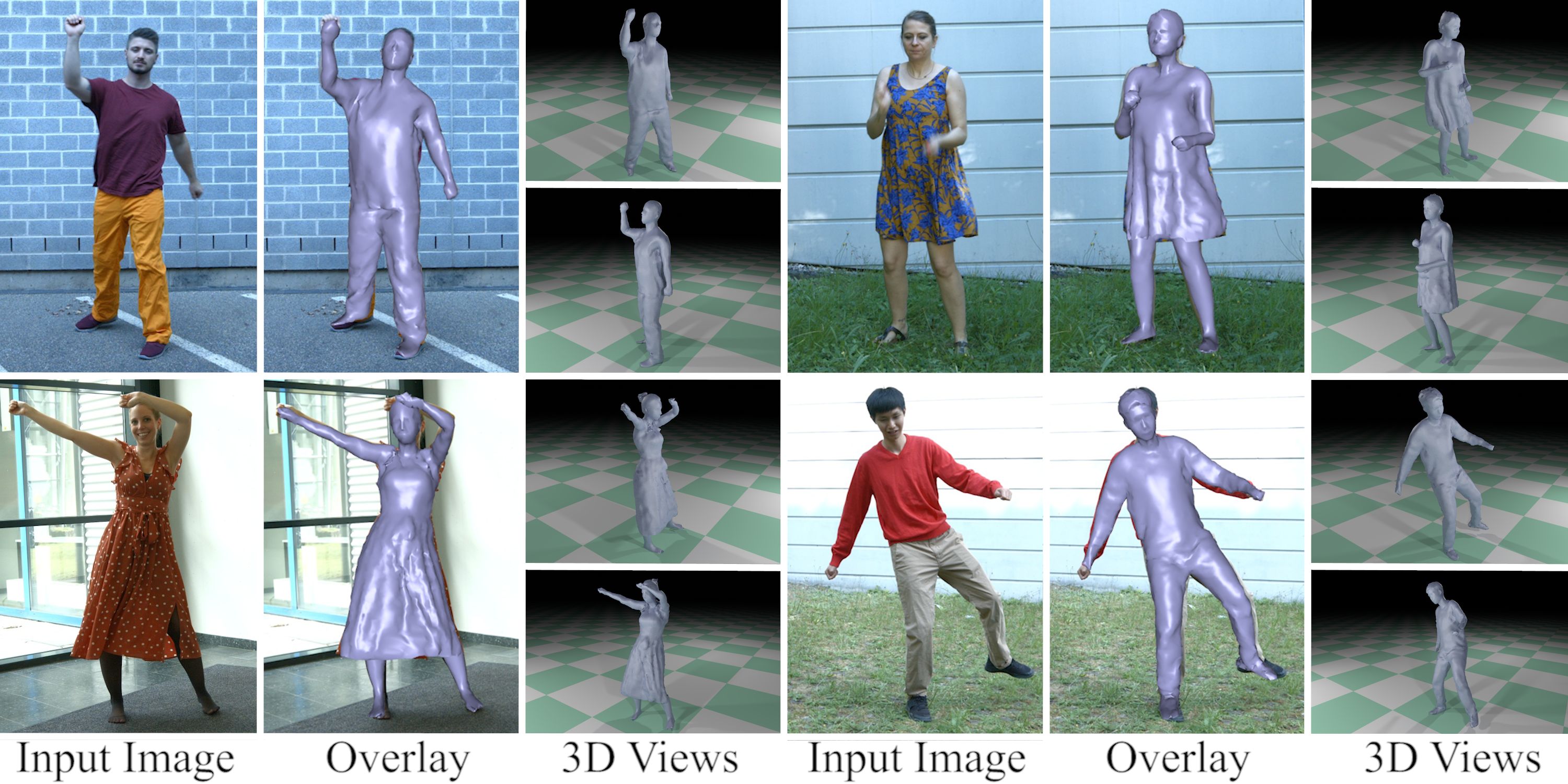}
	\end{center}
	\caption
	{
		We present the first learning-based approach for dense monocular human performance capture using weak multi-view supervision that not only predicts the pose but also the space-time coherent non-rigid deformations of the model surface.
	}
	\label{fig:teaser}
\end{figure}
%
%
%
\par
To overcome the shortcomings of multi-view approaches and with the advent of deep learning, some recent approaches~\cite{saito19,zheng19,alldieck19,alldieck18a,bhatnagar2019mgn, patel20, ma20} focus on predicting the 3D clothed surface of a human from a single color image.
In particular, some works leverage implicit surface representations such as 3D voxel grids~\cite{gabeur2019moulding,zheng19} or pixel aligned implicit surface functions~\cite{saito19,saito2020pifuhd}.
While such representations can handle topological surface changes as well as the capture of finer geometric details, they suffer from artifacts like missing limbs as the regressed deformations are not explicitly constrained by a discrete surface mesh.
Further, these methods are designed to reconstruct a surface per image.
Thus, naively applying those techniques to video streams results in individual geometries per frame which are not coherent over time.
To this end, another line of work~\cite{alldieck19,bhatnagar2019mgn,alldieck2019tex2shape,Pumarola_2019_ICCV} regresses surface deformation with respect to a naked human body model.
Here, coherence over time can be ensured and those works do not suffer from missing limbs since the underlying body model constrains the surface deformation.
However, one drawback of these approaches is that they do not capture the motion and surface deformations over time.
%
%
%
\par
The state-of-the-art human performance capture methods MonoPerfCap~\cite{xu18} and LiveCap~\cite{habermann19} densely track the deforming surface over time.
In contrast to the proposed method, they only predict sparse 2D/3D skeletal keypoints from the images and then perform an expensive optimization based pose and surface fitting.
By design, their method suffers from the monocular setting as the deformations can only be constrained by the single view and their pose results suffer from leaning forward artifacts which originates from their biased keypoint detections.
In contrast, we propose the first learning-based approach that jointly regresses the skeletal pose as well as the non-rigid surface deformation within a single inference pass resulting in a higher accuracy in terms of 3D surface and pose tracking as well as an improved robustness.
Specifically, two CNN-based models predict the skeletal joint angles and embedded deformation parameters of a differentiable mesh-based character representation from a single image.
By using an explicit mesh representation, our method has the advantage that the surface can be tracked over time which is key-essential for texturing and rendering in graphics.
Furthermore, the coarse-to-fine modeling of articulation and surface deformation ensures that the output of our method does not suffer from missing limbs, even when those are occluded or when the actor performs out-of-plane motions.
%
%
\par
In contrast to previous work~\cite{saito19,zheng19,alldieck19,bhatnagar2019mgn}, our method does not require any form of 3D supervision for training but instead leverages weak supervision in the form of multi-view videos which can be potentially also sparse.
To achieve this, we propose differentiable 3D-to-2D modules which allow us to train our deep architectures in an analysis-by-synthesis manner without using 3D ground truth.
At the core, our method leverages a person-specific 3D template of the actor as well as a multi-view video showing the actor while he performs a wide range of motions.
Then, our dedicated network modules predict the pose and surface deformation parameters that allow to pose and deform the template.
This deformed and posed template is then compared against sparse and dense multi-view observations extracted from the multi-view images in a differentiable manner.
Importantly, at test time our method only takes a single image as input and predicts the posed and non-rigidly deformed surface of the entire human including also the clothing.
In summary, the main technical contributions of our work are:
\begin{itemize}
	\item{A learning-based 3D human performance capture approach that jointly tracks the skeletal pose and the non-rigid surface deformations from monocular images.}
	\item{A new differentiable representation of deforming human surfaces which enables training from multi-view video footage directly.}
\end{itemize}
Our new model achieves high quality dense human performance capture results on our new challenging dataset, demonstrating, qualitatively and quantitatively, the advantages of our approach over previous work.
We experimentally show that our method produces reconstructions of higher accuracy and 3D stability, in particular in depth, than related work, also under difficult poses. 
%
%
\par
This work is an extended version of DeepCap~\cite{habermann20} where additional explanations, evaluations, comparisons, applications, and limitations are provided.
In particular, the character modeling and processing as well as the specific non-trivial training strategies are explained in more detail.
Further, we provide additional evaluations on all 4 subjects of the dataset of \cite{habermann20} and qualitative ablation results.
Last, we showcase potential applications and discuss limitations.

\section{Related Work} 
\label{sec:relatedwork}
In the following, we focus on related work in the field of dense 3D human performance capture and do not review work on sparse 2D pose estimation.
%
%
\par \noindent\textbf{Capture using Parametric Models.}
Monocular human performance capture is an ill-posed problem due to its high dimensionality and ambiguity.
Low-dimensional parametric models can be employed as shape and deformation prior.
First, model-based approaches leverage a set of simple geometric primitives \cite{plankers01, sminchisescu03b, sigal04, metaxas93}.
Recent methods employ detailed statistical models learned from thousands of high-quality 3D scans \cite{anguelov05, hasler10, park08, pons15, loper15, kadlecek16, meekyoung17, weiss11, helten13, zhang14a, bogo15}.
Deep learning is widely used to obtain 2D and/or 3D joint detections or 3D vertex positions that can be used to inform model fitting \cite{yinghao17,lassner17, mehta17, bogo16, kolotouros19}.
An alternative is to regress model parameters directly \cite{kanazawa18, pavlakos18, kanazawa19}.
Beyond body shape and pose, recent models also include facial expressions and hand motion \cite{pavlakos19, xiang18, joo18, romero17} leading to very expressive reconstruction results.
Recently, Zhou et al.~\cite{zhou2021monocular} also predict facial albedo and lighting parameters while even achieving real-time performance.
Since parametric body models do not represent garments, variation in clothing cannot be reconstructed, and therefore many methods recover the naked body shape under clothing \cite{balan07a, bualan08, zhang17, yang16}.
The full geometry of the actor can be reconstructed by non-rigidly deforming the base parametric model to better fit the observations \cite{rhodin16b, alldieck18a, alldieck18b, xiangdonglai2020}.
But they can only model tight clothes such as T-shirts and pants, but not loose apparel which has a different topology than the body model, such as skirts.
To overcome this problem, ClothCap \cite{pons17} captures the body and clothing separately, but requires active multi-view setups.
Physics based simulations have recently been leveraged to constrain the surface tracking~\cite{tao19, li2020deep} as well as the pose estimation~\cite{PhysCapTOG2020,PhysAwareTOG2021}, or to learn a model of clothing on top of SMPL (TailorNet~\cite{patel20}).
Instead, our method is based on person-specific templates including clothes and employs deep learning to predict clothing deformation based on monocular video directly.
%
%
\par \noindent\textbf{Depth-based Template-free Capture.}
Most approaches based on parametric models ignore clothing.
The other side of the spectrum are prior-free approaches based on one or multiple depth sensors.
Capturing general non-rigidly deforming scenes \cite{slavcheva17,guo17}, even at real-time frame rates \cite{newcombe15, innmann2016, guo17}, is feasible, but only works reliably for small, controlled, and slow motions.
Higher robustness can be achieved by using higher frame rate sensors \cite{guo18, kowdle18} or multi-view setups \cite{ye12, dou16, orts16, dou17, zhang14b}.
Techniques that are specifically tailored to humans increase robustness \cite{yu17, yu18, ye14} by integrating a skeletal motion prior \cite{yu17} or a parametric model \cite{yu18, wei12}.
HybridFusion~\cite{zheng18} additionally incorporates a sparse set of inertial measurement units.
These fusion-style volumetric capture techniques \cite{huang16, allain15, leroy17, collet15, prada17} achieve impressive results, but do not establish a set of dense correspondences between all frames.
In addition, such depth-based methods do not directly generalize to our monocular setting, have a high power consumption, and typically do not work well under sunlight.
%
%
\par \noindent\textbf{Monocular Template-free Capture.}
Quite recently, fueled by the progress in deep learning, many template-free monocular reconstruction approaches have been proposed.
Due to their regular structure, many implicit reconstruction techniques \cite{varol18, zheng19} make use of uniform voxel grids.
DeepHuman~\cite{zheng19} combines a coarse scale volumetric reconstruction with a refinement network to add high-frequency details.
Multi-view CNNs can map 2D images to 3D volumetric fields enabling reconstruction of a clothed human body at arbitrary resolution \cite{zeng18}.
SiCloPe~\cite{natsume18} reconstructs a complete textured 3D model, including cloth, from a single image.
PIFu~\cite{saito19}, its follow-up work~\cite{saito2020pifuhd}, and a real-time variant~\cite{li2020monocular} regress an implicit surface representation that locally aligns pixels with the global context of the corresponding 3D object.
Similar to our work, ARCH~\cite{Huang:ARCH:2020} proposes to evaluate the implicit surface function in pose canonical space which makes the learning task easier as the pose and surface deformation can be separated.
Unlike voxel-based representations, this implicit per-pixel representation is more memory efficient.
These approaches have not been demonstrated to generalize well to strong articulation.
Furthermore, implicit approaches do not recover frame-to-frame correspondences which are of paramount importance for downstream applications, e.g., in augmented reality and video editing.
In contrast, our method is based on a mesh representation and can explicitly obtain the per-vertex correspondences over time while being slightly less general.
%
%
\par \noindent\textbf{Template-based Capture.}
An interesting trade-off between being template-free and relying on parametric models are approaches that only employ a template mesh as prior.
Historically, template-based human performance capture techniques exploit multi-view geometry to track the motion of a person \cite{starck07}.
Some systems also jointly reconstruct and obtain a foreground segmentation \cite{bray06, brox10, liu11, wu12}.
Given a sufficient number of multi-view images as input, some approaches \cite{carranza03, cagniart10, de08} align a personalized template model to the observations using non-rigid registration.
All the aforementioned methods require expensive multi-view setups and are not practical for consumer use.
Depth-based techniques enable template tracking from less cameras \cite{zollhoefer2014, ye12} and reduced motion models \cite{wu13, gall09, vlasic08, liu11} increase tracking robustness.
Recently, capturing 3D dense human body deformation just with a single RGB camera has been enabled \cite{xu18} and real-time performance has been achieved \cite{habermann19}.
However, their methods rely on expensive optimization leading either to very long per-frame computation times \cite{xu18} or the need for two graphics cards \cite{habermann19}.
Similar to them, our approach also employs a person-specific template mesh.
But differently, our method directly learns to predict the skeletal pose and the non-rigid surface deformations.
As shown by our experimental results, benefiting from our multi-view based self-supervision, our reconstruction accuracy significantly outperforms the existing methods.
\section{Method}
\label{sec:method}
%
%
\begin{figure}[t]
	\begin{center}
		\includegraphics[width=\linewidth]{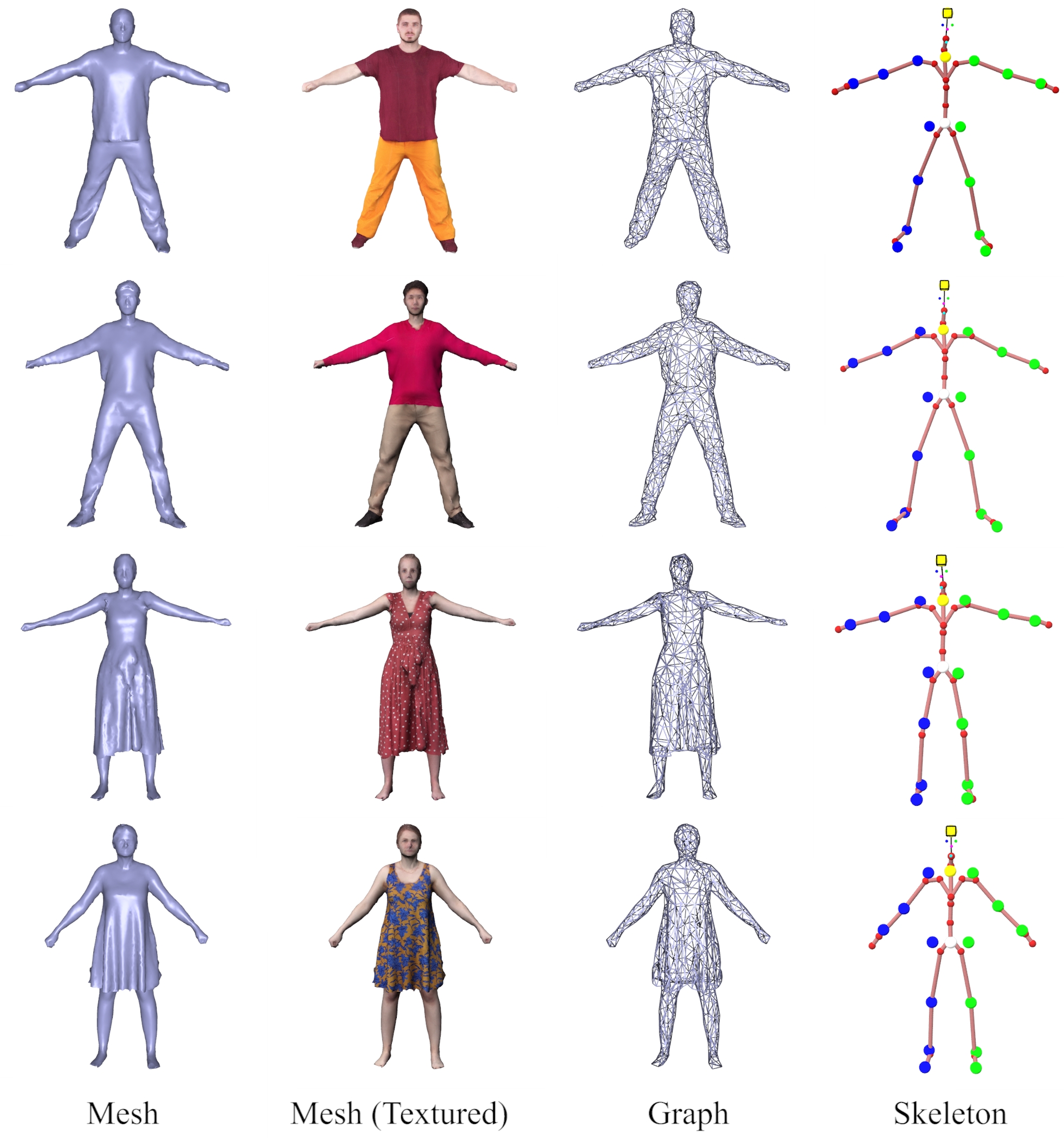}
	\end{center}
	\caption
	{
		Character models. Here, we show the character model of $S1$ to $S4$ (top to bottom) of our new dataset. It consists of the textured mesh, the underlying embedded deformation graph as well as the attached skeleton.
	}
	\label{fig:template}
\end{figure}
\begin{figure*}[t]
	    \centering
	    \includegraphics[width=\textwidth]{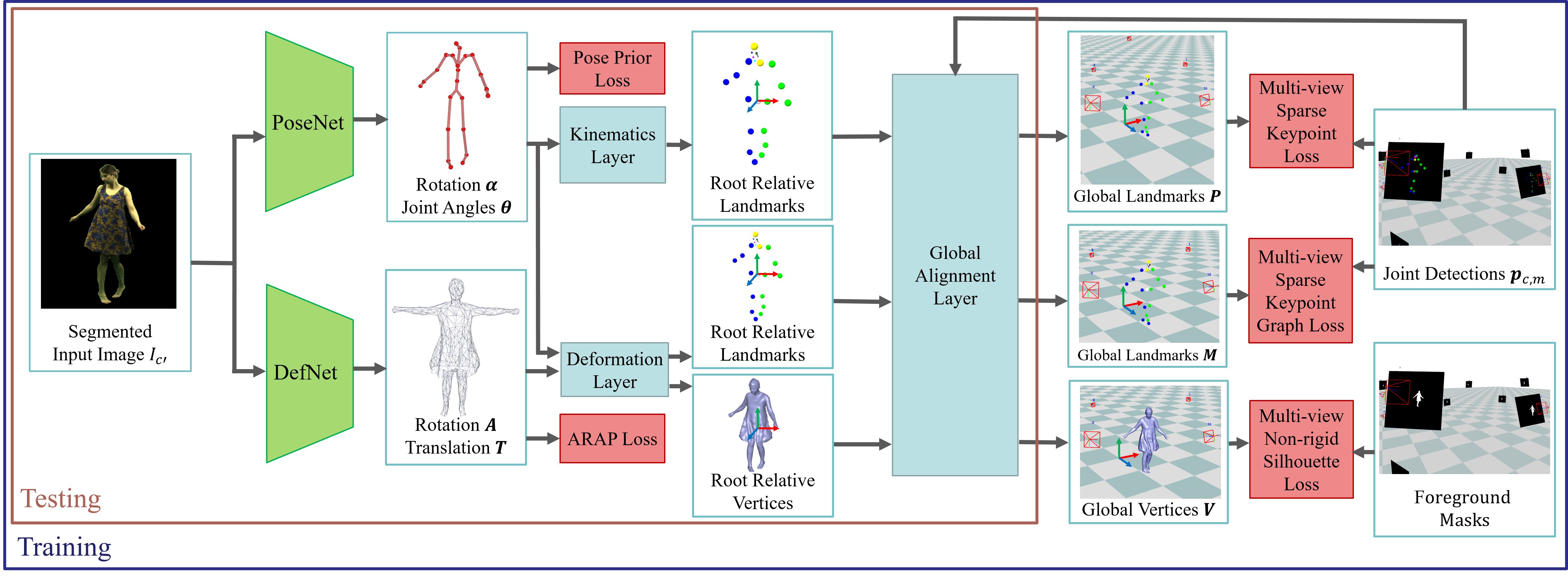} 
	    \caption
	    {
	    	Overview of our approach. 
	    	Our method takes a single segmented image as input. 
	    	First, our pose network, \textit{PoseNet}, is trained to predict the joint angles and the camera relative rotation using sparse multi-view 2D joint detections as weak supervision. 
	    	Second, the deformation network, \textit{DefNet}, is trained to regress embedded graph rotation and translation parameters to account for non-rigid deformations. 
	    	To train DefNet, multi-view 2D joint detections and silhouettes are used for supervision. 
    	}
	    \label{fig:overview}
    	\vspace{-12pt}
\end{figure*}
%
%
%
\par 
Given a single RGB video of a moving human in general clothing, our goal is to capture the dense deforming surface of the full body.
This is achieved by training a neural network consisting of two components:
As illustrated in Fig.~\ref{fig:overview}, our pose network, \textit{PoseNet}, estimates the skeletal pose of the actor in the form of joint angles from a monocular image (Sec.~\ref{sec:poseNetwork}). 
Next, our deformation network, \textit{DefNet}, regresses the non-rigid deformation of the dense surface, which cannot be modeled by the skeletal motion, in the embedded deformation graph representation (Sec.~\ref{sec:deformationNetwork}).
To avoid generating dense 3D ground truth annotation, our network is trained in a weakly supervised manner.
To this end, we propose a fully differentiable human deformation and rendering model, which allows us to compare the rendering of the human body model to the 2D image evidence and back-propagate the losses.
For training, we first capture a video sequence in a calibrated multi-camera green screen studio (Sec.~\ref{sec:templateDataAndAcquisition}).
Note that our multi-view video is only used during training.
More details about training are provided in Sec.~\ref{sec:implementationDetails}.
At test time we only require a single RGB video to perform dense non-rigid tracking.
%
%
\subsection{Template and Data Acquisition}
\label{sec:templateDataAndAcquisition}
%
%
\par \noindent\textbf{Character Model.}
Our method relies on a person-specific 3D template model.
To this end, we first scan the actor with a 3D scanner~\cite{treedys} to obtain the textured mesh (see Fig.~\ref{fig:template}).
Next, we rig a skeleton onto the template which consists of 23 joints and 21 attached landmarks (17 body and 4 face landmarks) and it is parameterized with 27 joint angles $\boldsymbol{\theta} \in \mathbb{R}^{27}$, the camera relative rotation $\boldsymbol{\alpha} \in \mathbb{R}^{3}$ and translation $\mathbf{t} \in \mathbb{R}^{3}$. 
The landmark placement follows the convention of OpenPose~\cite{cao17,cao18,simon17,wei16}.
To model the non-rigid surface deformation, we automatically build an embedded deformation graph $\mathcal{G}$ with $K$ nodes.
The connections of a node $k$ to neighboring nodes are denoted as the set $\mathcal{N}_\mathrm{n}(k)$.
The position of the graph nodes is denoted as $\mathbf{G}\in \mathbb{R}^{K \times 3}$ where $\mathbf{G}_k$ is the position of node $k$.
Finally, the vertex-to-node weights between the graph node $k$ and the template vertex $i$ are defined as $w_{i,k}$ and $\mathcal{N}_{\mathrm{vn}}(i)$ denotes the set of nodes that influence vertex $i$.
The nodes are parameterized with Euler angles $\mathbf{A} \in \mathbb{R}^{K \times 3}$ and translations $\mathbf{T} \in \mathbb{R}^{K \times 3}$.
Similar to \cite{habermann19}, we segment the mesh into different non-rigidity classes resulting in per-vertex rigidity weights $s_{i}$.
This allows us to model varying deformation behaviors of different surface materials, e.g. skin deforms less than clothing (see Eq.~\ref{eq:lossARAP}).
%
%
\par \noindent\textbf{Training Data.}
To acquire the training data, we record a multi-view video using $C$ calibrated cameras of the actor doing various actions in a calibrated multi-camera studio with green screen.
To provide weak supervision for the training, we first perform 2D pose detection on the sequences using OpenPose~\cite{cao17,cao18,simon17,wei16} and apply temporal filtering.
Then, we generate the foreground mask using color keying and compute the corresponding distance transform image $D_{f,c}$~\cite{borgefors86}, where $f \in [0,F]$ and $c \in [0,C]$ denote the frame index and camera index, respectively.
During training, we randomly sample one camera view $c'$ and frame $f'$ for which we crop the recorded image with a bounding box, based on the 2D joint detections.
The final training input image $I_{f',c'} \in \mathbb{R}^{256 \times 256 \times 3}$ is obtained by removing the background and augmenting the foreground with random brightness, hue, contrast and saturation changes.
For simplicity, we describe the operation on frame $f'$ and omit the subscript $f'$ in following equations.
%
%
%
\subsection{Pose Network}
\label{sec:poseNetwork}
In our \textit{PoseNet}, we use ResNet50~\cite{he16} pretrained on ImageNet~\cite{deng09} as backbone and modify the last fully connected layer to output a vector containing the joint angles $\boldsymbol{\theta}$ and the camera relative root rotation $\boldsymbol{\alpha}$, given the input image $I_{c'}$. 
Since generating the ground truth for $\boldsymbol{\theta}$ and $\boldsymbol{\alpha}$ is a non-trivial task, we propose weakly supervised training based on fitting the skeleton to multi-view 2D joint detections.
%
%
\par \noindent\textbf{Kinematics Layer.}
To this end, we introduce a kinematics layer as the differentiable function that takes the joint angles $\boldsymbol{\theta}$ and the camera relative rotation $\boldsymbol{\alpha}$ and computes the positions $\mathbf{P}_{c'} \in \mathbb{R}^{M \times 3}$ of the $M$ 3D landmarks attached to the skeleton (17 body joints and 4 face landmarks). 
Note that $\mathbf{P}_{c'}$ lives in a camera-root-relative coordinate system.
In order to project the landmarks to other camera views, we need to transform $\mathbf{P}_{c'}$ to the world coordinate system:
\begin{equation}
\label{eq:global_transform}
    \mathbf{P}_m = \mathbf{R}_{c'}^{T} \mathbf{P}_{c',m} + \mathbf{t},
\end{equation}
where $\mathbf{R}_{c'}$ is the rotation matrix of the input camera $c'$ and $\mathbf{t}$ is the global translation of the skeleton.
%
%
\par \noindent\textbf{Global Alignment Layer.}
To obtain the global translation $\mathbf{t}$, we propose a global alignment layer that is attached to the kinematics layer.
It localizes our skeleton model in the world space, such that the globally rotated landmarks $\mathbf{R}_{c'}^{T} \mathbf{P}_{c',m}$ project onto the corresponding detections in all camera views.
This is done by minimizing the distance between the rotated landmarks $\mathbf{R}_{c'}^{T} \mathbf{P}_{c',m}$ and the corresponding rays cast from the camera origin $\mathbf{o}_c$ to the 2D joint detections:
\begin{equation} \label{eq:EGlobalAlign}
\sum_c \sum_m
\sigma_{c,m}
		\lVert
			(\mathbf{R}_{c'}^T \mathbf{P}_{c',m} + \mathbf{t} -\mathbf{o}_c) \times \mathbf{d}_{c,m}
		\rVert^2,
\end{equation}
where $\mathbf{d}_{c,m}$ is the direction of a ray from camera $c$ to the 2D joint detection $\mathbf{p}_{c,m}$ corresponding to landmark $m$:
\begin{equation}
\mathbf{d}_{c,m}= 
\frac
	{(\mathbf{E}_c^{-1} \tilde{\mathbf{p}}_{c,m})_{xyz} -\mathbf{o}_c }
	{\lVert(\mathbf{E}_c^{-1} \tilde{\mathbf{p}}_{c,m})_{xyz} -\mathbf{o}_c \rVert}
	\mathrm{.}
\end{equation}
Here, $\mathbf{E}_c \in \mathbb{R}^{4 \times 4}$ is the projection matrix of camera $c$ and $\tilde{\mathbf{p}}_{c,m} = (\mathbf{p}_{c,m},1,1)^T$.
Each point-to-line distance is weighted by the joint detection confidence $\sigma_{c,m}$, which is set to zero if below $0.4$. 
The minimization problem of Eq.~\ref{eq:EGlobalAlign} can be solved in closed form:
\begin{equation}\label{eq:globalalignop}
    \mathbf{t} = 
    \mathbf{W}^{-1}
	    \sum_{c,m}
	    	\mathbf{D}_{c,m}(\mathbf{R}_{c'}^T \mathbf{P}_{c',m} -\mathbf{o}_{c}) + \mathbf{o}_{c} - \mathbf{R}_{c'}^T \mathbf{P}_{c',m} 
	\mathrm{,}
\end{equation}
where 
\begin{equation}
\mathbf{W}
=
	\sum_c
	\sum_m 
	\mathbf{I}-\mathbf{D}_{c,m}
	\mathrm{.}
\end{equation}
Here, $\mathbf{I}$ is the $3\times3$ identity matrix and $\mathbf{D}_{c,m}=\mathbf{d}_{c,m}\mathbf{d}_{c,m}^T$.
Note that the operation in Eq.~\ref{eq:globalalignop} is differentiable with respect to the landmark position $\mathbf{P}_{c'}$.
%
%
\par \noindent\textbf{Sparse Keypoint Loss.}
Our 2D sparse keypoint loss for the \textit{PoseNet} can be expressed as
\begin{equation} \label{eq:lossKeypoint}
	\mathcal{L}_{\mathrm{kp}}(\mathbf{P}) = 
	\sum_c \sum_m
				\lambda_{m} \sigma_{c,m}
				\lVert 
					\pi_c\left(	
						\mathbf{P}_m
					\right)
					-
					\mathbf{p}_{c,m}
				\rVert^2 \mathrm{,}
\end{equation}
which ensures that each landmark projects onto the corresponding 2D joint detections $\mathbf{p}_{c,m}$ in all camera views.
Here, $\pi_c$ is the projection function of camera $c$ and $\sigma_{c,m}$ is the same as in Eq.~\ref{eq:EGlobalAlign}.
$\lambda_{m}$ is a hierarchical re-weighting factor that varies during training for better convergence. 
More precisely, for the first one third of the training iterations per training stage (see Sec.~\ref{sec:implementationDetails}) for \textit{PoseNet}, we multiply the keypoint loss with a factor of $\lambda_{m}=3$ for all torso markers and with a factor of $\lambda_{m}=2$ for elbow and knee markers. 
For all other markers, we set $\lambda_{m}=1$.
For the remaining iterations, we set $\lambda_{m}=3$ for all markers.
This re-weighting allows us to let the model first focus on the global rotation (by weighting torso markers higher than others). 
We found that this gives better convergence during training and joint angles overshoot less often, especially at the beginning of training.
%
%
\par \noindent\textbf{Pose Prior Loss.}
To avoid unnatural poses, we impose a pose prior loss on the joint angles
\begin{equation} \label{eq:limit}
	\mathcal{L}_\mathrm{limit}(\boldsymbol{\theta}) = \sum_{i=1}^{27}{ \Psi( \boldsymbol{\theta}_i ) }
\end{equation}
\begin{equation} \label{eq:psi}
	\Psi(x)
	=
	\begin{cases}
		(x - \boldsymbol{\theta}_{\mathrm{max},i})^2,\text{ if } x > \boldsymbol{\theta}_{\mathrm{max},i}\\
		(\boldsymbol{\theta}_{\mathrm{min},i} - x)^2 \, ,\text{ if } x < \boldsymbol{\theta}_{\mathrm{min},i}\\
		0 \qquad \qquad \; \; \; , \text{ otherwise}
	\end{cases}\mathrm{,}
\end{equation}
that encourages that each joint angle $\boldsymbol{\theta}_i$ stays in a range $[\boldsymbol{\theta}_{\mathrm{min},i},\boldsymbol{\theta}_{\mathrm{max},i}]$ depending on the anatomic constraints.
%
%
%

\subsection{Deformation Network}
\label{sec:deformationNetwork}
%
%
With the skeletal pose from \textit{PoseNet} alone, the non-rigid deformation of the skin and clothes cannot be fully explained.
Therefore, we disentangle the non-rigid deformation and the articulated skeletal motion.
\textit{DefNet} takes the input image $I_{c'}$ and regresses the non-rigid deformation parameterized with rotation angles $\mathbf{A}$ and translation vectors $\mathbf{T}$ of the nodes of the embedded deformation graph. 
\textit{DefNet} uses the same backbone architecture as \textit{PoseNet}, while the last fully connected layer outputs a $6K$-dimensional vector reshaped to match the dimensions of $\mathbf{A}$ and $\mathbf{T}$.
The weights of \textit{PoseNet} are fixed while training \textit{DefNet}.
Again, we do not use direct supervision on $\mathbf{A}$ and $\mathbf{T}$.
Instead, we propose a deformation layer with differentiable rendering and use multi-view silhouette-based weak supervision.
%
%
\par \noindent\textbf{Deformation Layer.}
The deformation layer takes $\mathbf{A}$ and $\mathbf{T}$ from \textit{DefNet} as input to non-rigidly deform the surface
\begin{equation} \label{eq:vnr}
	\mathbf{Y}_{i}= 
	\sum_{k \in \mathcal{N}_{\mathrm{vn}}(i)}
	w_{i,k}
	(
		R(\mathbf{A}_k)(\hat{\mathbf{V}}_i-\mathbf{G}_k) +\mathbf{G}_k + \mathbf{T}_k
	)
	\mathrm{.}
\end{equation}
Here, $\mathbf{Y}, \hat{\mathbf{V}} \in \mathbb{R}^{N \times 3}$ are the vertex positions of the deformed and undeformed template mesh, respectively.
$w_{i,k}$ are vertex-to-node weights, but in contrast to \cite{sumner07} we compute them based on geodesic distances. 
$\mathbf{G}\in \mathbb{R}^{K \times 3}$ are the node positions of the undeformed graph, $\mathcal{N}_{\mathrm{vn}}(i)$ is the set of nodes that influence vertex $i$, and $R(\cdot)$ is a function that converts the Euler angles to rotation matrices.
We further apply the skeletal pose on the deformed mesh vertices to obtain the vertex positions in the input camera space
\begin{equation} \label{eq:vi}
	\mathbf{V}_{\mathrm{c'},i} =
	\sum_{k \in \mathcal{N}_{\mathrm{vn}}(i)}
	w_{i,k}
	(
	R_{\mathrm{sk},k}(\boldsymbol{\theta}, \boldsymbol{\alpha}) \mathbf{Y}_{i} + t_{\mathrm{sk},k}(\boldsymbol{\theta}, \boldsymbol{\alpha})
	)
	\mathrm{,}
\end{equation}
where the node rotation $R_{\mathrm{sk},k}$ and translation $t_{\mathrm{sk},k}$ are derived from the pose parameters using dual quaternion skinning~\cite{kavan07}.
Eq.~\ref{eq:vnr} and Eq.~\ref{eq:vi} are differentiable with respect to pose and graph parameters.
Thus, our layer can be integrated in the learning framework and gradients can be propagated to \textit{DefNet}.
So far, $\mathbf{V}_{\mathrm{c'},i}$ is still rotated relative to the camera $c'$ and located around the origin.
To bring them to global space, we apply the inverse camera rotation and the global translation, defined in Eq.~\ref{eq:globalalignop}, $\mathbf{V}_i = \mathbf{R}_{c'}^{T} \mathbf{V}_{\mathrm{c'},i} + \mathbf{t}$.
%
%
\par \noindent\textbf{Non-rigid Silhouette Loss.}
This loss encourages that the non-rigidly deformed mesh matches the multi-view silhouettes in all camera views.
It can be formulated using the distance transform representation~\cite{borgefors86}
\begin{equation} \label{eq:lossSil}
	\mathcal{L}_{\mathrm{sil}}(	\mathbf{V}) = 
		\sum_{c} \sum_{i \in \mathcal{B}_c} 
				\rho_{c,i}
				\|D_{c}\left(
						\pi_c\left(	
							\mathbf{V}_i
						\right)
					\right)
			 	\|^2\mathrm{.}
\end{equation}
Here, $\mathcal{B}_c $ is the set of vertices that lie on the boundary when the deformed 3D mesh is projected onto the distance transform image $D_{c}$ of camera $c$.
Those vertices are computed by rendering a depth map using a custom CUDA-based rasterizer that can be easily integrated into deep learning architectures as a separate layer.
The vertices that project onto a depth discontinuity (background vs. foreground) in the depth map are treated as boundary vertices.
$\rho_{c,i}$ is a directional weighting~\cite{habermann19} that guides the gradient in $D_{c}$.
The silhouette loss ensures that the boundary vertices project onto the zero-set of the distance transform, \textit{i.e.}, the foreground silhouette.
%
%
\par \noindent\textbf{Sparse Keypoint Graph Loss.}
Only using the silhouette loss can lead to wrong mesh-to-image assignments, especially for highly articulated motions.
To this end, we use a sparse keypoint loss to constrain the mesh deformation, which is similar to the keypoint loss for \textit{PoseNet} in Eq.~\ref{eq:lossKeypoint}
\begin{multline} \label{eq:keypointGraphLoss}
\mathcal{L}_{\mathrm{kpg}}(\mathbf{M}) = 
\sum_c \sum_m
\sigma_{c,m}
\lVert 
\pi_c\left(	
\mathbf{M}_m
\right)
-
\mathbf{p}_{c,m}
\rVert^2
\mathrm{.}
\end{multline}
Differently from Eq.~\ref{eq:lossKeypoint}, the deformed and posed landmarks $\mathbf{M}$ are derived from the embedded deformation graph.
To this end, we can deform and pose the canonical landmark positions by attaching them to its closest graph node $g$ in canonical pose with weight $w_{m,g}=1.0$. 
Landmarks can then be deformed according to Eq.~\ref{eq:vnr}, ~\ref{eq:vi}, resulting in $\mathbf{M}_{c'}$ which is brought to global space via $\mathbf{M}_m = \mathbf{R}_{c'}^{T} \mathbf{M}_{\mathrm{c'},m} + \mathbf{t}$.
%
%
\par \noindent\textbf{As-rigid-as-possible Loss.}
To enforce local smoothness of the surface, we impose an as-rigid-as-possible loss~\cite{sorkine07}
\begin{equation} \label{eq:lossARAP}
\mathcal{L}_\mathrm{arap}(\mathbf{A},\mathbf{T}) = 
\sum_k \sum_{l \in \mathcal{N}_\mathrm{n}(k)}
u_{k,l}
\lVert 
d_{k,l}(\mathbf{A},\mathbf{T}) 
\rVert_1,
\end{equation}
where
$$
d_{k,l}(\mathbf{A},\mathbf{T})\! \!= \!\!
R(\mathbf{A}_k) (\mathbf{G}_l - \mathbf{G}_k) + \mathbf{T}_k + \mathbf{G}_k
- (\mathbf{G}_l + \mathbf{T}_l).
$$
$\mathcal{N}_\mathrm{n}(k)$ is the set of indices of the neighbors of node $k$.
In contrast to \cite{sorkine07}, we propose weighting factors $u_{k,l}$ that influence the rigidity of respective parts of the graph.
We derive $u_{k,l}$ by averaging all per-vertex rigidity weights $s_{i}$ \cite{habermann19} for all vertices (see Sec.~\ref{sec:templateDataAndAcquisition}), which are connected to node $k$ or $l$.
Thus, the mesh can deform either less or more depending on the surface material. 
For example, graph nodes that are mostly connected to vertices on a skirt can deform more freely than nodes that are mainly connected to vertices on the skin.
Without this loss, the deformations can strongly drift along the visual hull carved by the silhouette images without receiving any penalty leading to strong visual artifacts.
\subsection{In-the-wild Domain Adaptation}
\label{sec:domainAdaptation}
Since our training set is captured in a green screen studio and our test set is captured in the wild, there is a significant domain gap between them, due to different lighting conditions and camera response functions.
To improve the performance of our method on in-the-wild images, we fine-tune our networks on the monocular test images for a small number of iterations using the same 2D keypoint and silhouette losses as before, \emph{but only on a single view}.
This drastically improves the performance at test time as shown in the supplemental material.
\section{Implementation Details}
\label{sec:implementationDetails}
Both network architectures as well as the GPU-based custom layers are implemented in the Tensorflow framework~\cite{tensorflow}.
We use the Adam optimizer~\cite{kingma14} in all our experiments.
%
%
\par \noindent \textbf{Template Acquisition and Rigging.}
To create the textured mesh (see Fig.~\ref{fig:template}), we capture the person in a static T-pose with an RGB-based scanner\footnote{\url{https://www.treedys.com/}} which has 134 RGB cameras resulting in 134 images $\mathcal{I}_{\mathrm{rec}} = \{I_{\mathrm{rec}_1}, \cdots, I_{\mathrm{rec}_{134}}\}$. 
The textured 3D geometry is obtained by leveraging a commercial 3D reconstruction software, called Agisoft Metashape\footnote{\url{http://www.agisoft.com}}, that takes as input the images $\mathcal{I}_{\mathrm{rec}}$ and reconstructs a textured 3D mesh of the person (see Fig.~\ref{fig:template}).
We apply Metashape's mesh simplification to reduce the number of vertices $N$ and Meshmixer's\footnote{\url{http://www.meshmixer.com/}} remeshing to obtain roughly uniform shaped triangular surfaces.
Next, we automatically fit the skeleton (see Fig.~\ref{fig:template}) to the 3D mesh by fitting the SMPL model~\cite{loper15}.
To this end, we first optimize the pose by performing a sparse non-rigid ICP where we use the head, hands and feet as feature points since they can be easily detected in a T-pose.
Then, we perform a dense non-rigid ICP on vertex level to obtain the final pose and shape parameters.
For clothing types that roughly follow the human body shape, e.g., pants and shirt, we propagate the per-vertex skinning weights of the naked SMPL model to the template vertices.
For other types of clothing, like skirts and dresses, we leverage Blenders's\footnote{\url{https://www.blender.org/}} automated skinning weight computation.
%
%
\par \noindent \textbf{Embedded Graph Construction.}
To build the embedded deformation graph $\mathcal{G}$ with $K$ the template mesh is decimated to around 500 vertices (see Fig.~\ref{fig:template}).
The connections of a node $k$ to neighboring nodes are given by the vertex connections of the decimated mesh.
For each vertex of the decimated mesh we search for the closest vertex on the template mesh in terms of Euclidean distance.
These points then define the position of the graph nodes.
To compute the vertex-to-node weights $w_{i,k}$, we measure the geodesic distance between the graph node $k$ and the template vertex $i$.
%
%
\par \noindent \textbf{Multi-view video.}
The number of frames per subject varies between 26000 and 38000 depending on how fast the person performed all the motions.
We used $C$ calibrated and synchronized cameras with a resolution of $1024\times1024$ for capturing where for all subjects we used between 11 and 14 cameras.
The original image resolution is too large to transfer all the distance transform images to the GPU during training.
Fortunately, most of the image information is anyways redundant since we are only interested in the image region where the person is.
Therefore, we crop the distance transform images using the bounding box that contains the segmentation mask with a conservative margin.
Finally, we resize it to a resolution of $350\times350$ without loosing important information.
%
%
\par \noindent \textbf{Training Strategy for PoseNet.}
As we are interested in joint angle regression, one has to note that multiple solutions for the joint angles exist due the fact that every correct solution can be multiplied by $2\pi$ leading to the same loss value. 
To this end, training has to be carefully designed. 
In general, our strategy first focuses on the torso markers by giving them more weight (see Sec.~\ref{sec:poseNetwork}).
Using this strategy, the global rotation will be roughly correct and joint angles are slowly trained to avoid overshooting of angular values.
This is further ensured by our limits term.
After several epochs, when the network already learned to fit the poses roughly, we turn off the regularization and let it refine the angles further.
More precisely, the training of \textit{PoseNet} proceeds in three stages.
First, we train \textit{PoseNet} for $120k$ iterations with a learning rate of $10^{-5}$ and weight $\mathcal{L}_{\mathrm{kp}}$ with $0.01$.
$\mathcal{L}_\mathrm{limit}$ has a weight of $1.0$ for the first $40k$ iterations. 
Between $40k$ and $60k$ iterations we re-weight $\mathcal{L}_\mathrm{limit}$ with a factor of $0.1$. 
Finally, we set $\mathcal{L}_\mathrm{limit}$ to zero for the remaining training steps.
Second, we train \textit{PoseNet} for another $120k$ iterations with a learning rate of $10^{-6}$ and $\mathcal{L}_{\mathrm{kp}}$ is weighted with a factor of $10^{-4}$.
Third, we train \textit{PoseNet} again $120k$ iterations with a learning rate of $10^{-6}$ and $\mathcal{L}_{\mathrm{kp}}$ is weighted with a factor of $10^{-5}$.
We always use a batch size of $90$.
%
%
\par \noindent \textbf{Training Strategy for DefNet.}
We train \textit{DefNet} for $120k$ iterations with a batch size of $50$.
We used a learning rate of $10^{-5}$ and weight $\mathcal{L}_{\mathrm{sil}}$, $\mathcal{L}_{\mathrm{kpg}}$, and $\mathcal{L}_{\mathrm{arap}}$ with $1k$, $0.05$, and $1.5k$, respectively.
%
%
\par \noindent \textbf{Training Strategy for the Domain Adaptation.}
To fine-tune the network for in-the-wild monocular test sequences, we train the pre-trained \textit{PoseNet} and \textit{DefNet} for 250 iterations, respectively.
To this end, we replace the multi-view losses with a single view loss which can be trivially achieved.
For \textit{PoseNet}, we disable $\mathcal{L}_{\mathrm{limit}}$ and weight $\mathcal{L}_{\mathrm{kp}}$ with $10^{-6}$.
For \textit{DefNet}, we weight $\mathcal{L}_{\mathrm{sil}}$, $\mathcal{L}_{\mathrm{kpg}}$, and $\mathcal{L}_{\mathrm{arap}}$ with $1k$, $0.05$, and $1.5k$ respectively.
Further, we use a learning rate of $10^{-6}$ and use the same batch sizes as before.
This fine-tuning in total takes around 5 minutes.
All hyperparameters are empirically determined and fixed across different subjects.
\section{Results}
\label{sec:evaluation}
\begin{figure*}[t]
	\begin{center}
		\includegraphics[width=\linewidth]{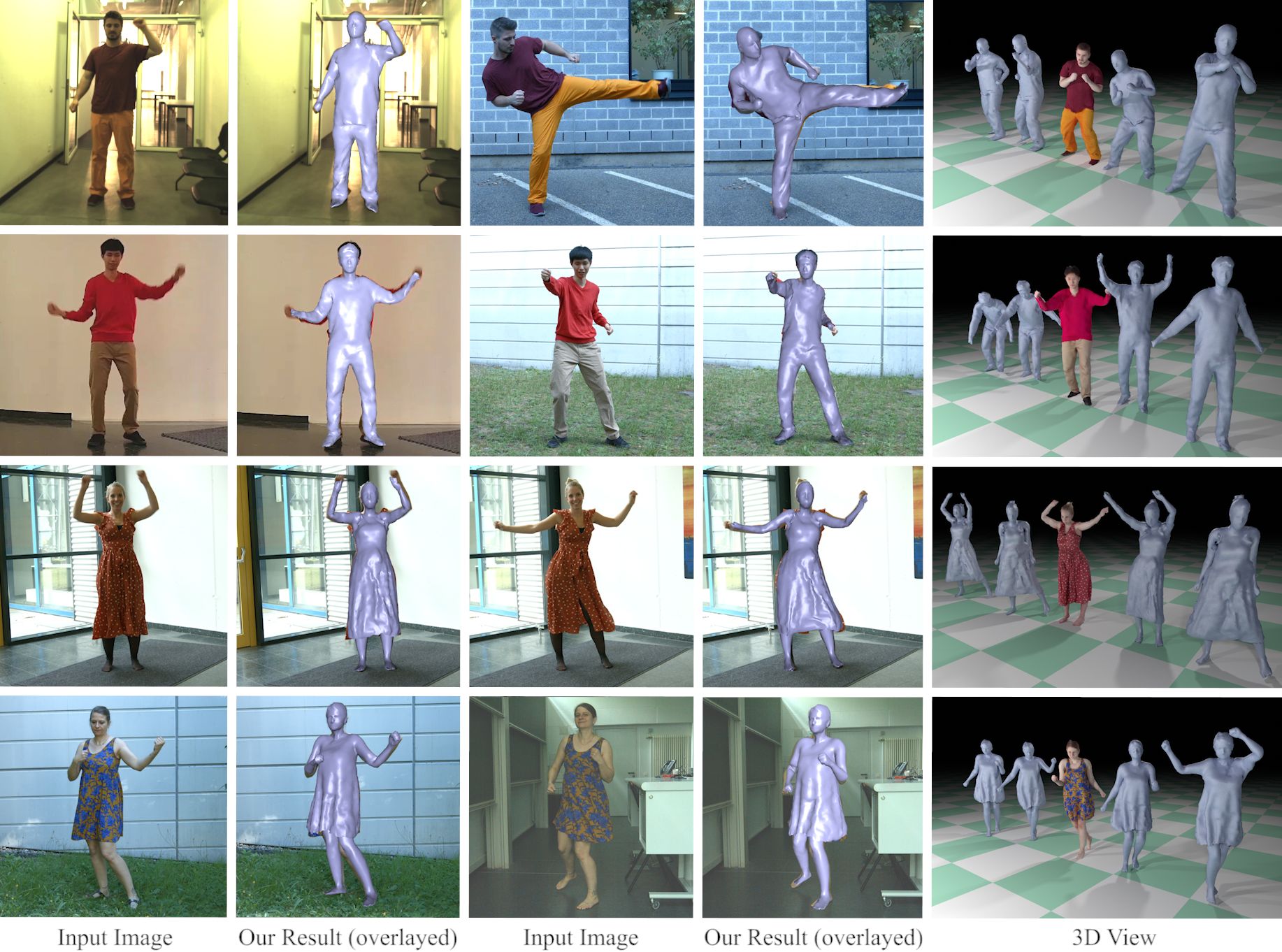}
	\end{center}
	\caption
	{
		Qualitative results. 
		Each row shows results for a different person with varying types of apparel. 
		We visualize input frames and our reconstruction overlayed to the corresponding frame.
		Note that our results precisely overlay to the input.
		Further, we show our reconstructions from a virtual 3D viewpoint.
		Note that they also look plausible in 3D.
	}
	\label{fig:qualitative}
\end{figure*}
All our experiments were performed on a machine with an NVIDIA Tesla V100 GPU.
A forward pass of our method takes around 50ms, which breaks down to 25ms for \textit{PoseNet} and 25ms for \textit{DefNet}.
During testing, we use the off-the-shelf video segmentation method of~\cite{caelles17} to remove the background in the input image.
Our method requires OpenPose's 2D joint detections~\cite{cao17,cao18,simon17,wei16} as input during testing to crop the frames and to obtain the 3D global translation with our global alignment layer.
Finally, we temporally smooth the output mesh vertices with a Gaussian kernel of size 5 frames.
%
%
\par
\begin{figure}[t]
	\begin{center}
		\includegraphics[width=\linewidth]{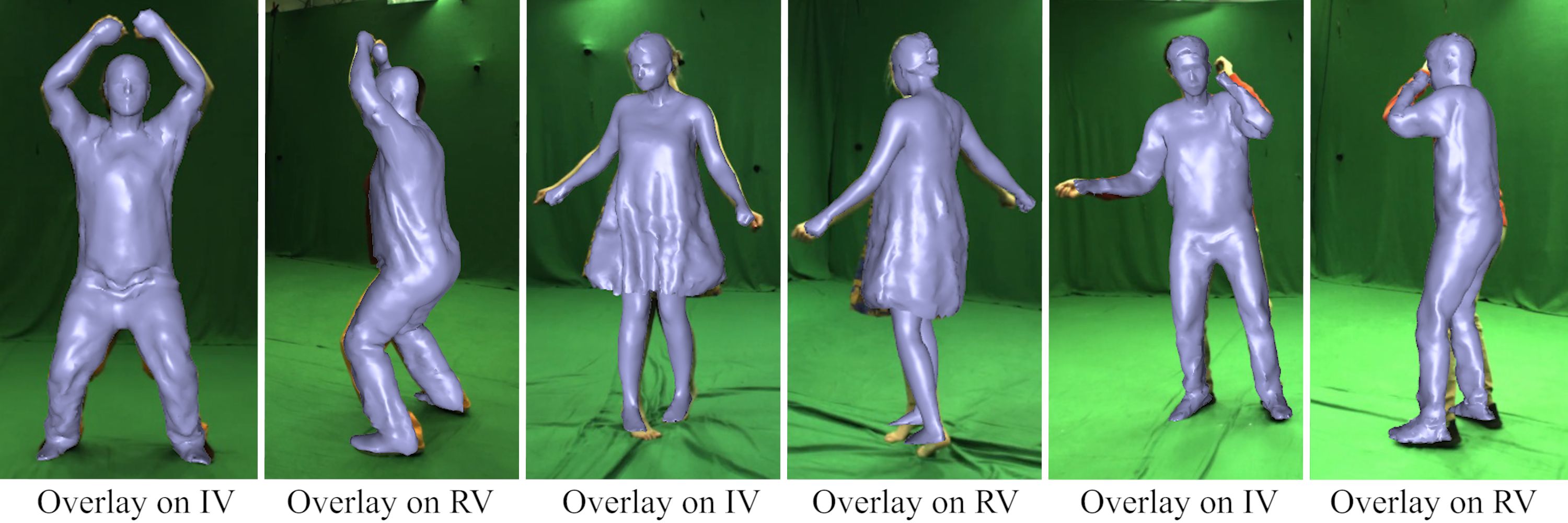}
	\end{center}
	\caption
	{
		Results on our evaluation sequences where input views (IV) and reference views (RV) are available.
		Note that our reconstruction also precisely overlays on RV even though they are not used for tracking.
	}
	\label{fig:referenceview}
\end{figure}

 \noindent \textbf{Dataset.}
We evaluate our approach on 4 subjects (\textit{S1} to \textit{S4}) with varying types of apparel.
For qualitative evaluation, we recorded 13 in-the-wild sequences in different indoor and outdoor environments shown in Fig.~\ref{fig:qualitative}.
For quantitative evaluation, we captured 4 sequences in a calibrated multi-camera green screen studio (see Fig.~\ref{fig:referenceview}), for which we computed the ground truth 3D joint locations using the multi-view motion capture software, The Captury~\cite{captury}, and we use a color keying algorithm for ground truth foreground segmentation.
All sequences contain a large variety of motions, ranging from simple ones like walking up to more difficult ones like fast dancing or baseball pitching. 
We will release the dataset for future research.
%
%
%
%
\par
\begin{figure}[t]
	\begin{center}
		\includegraphics[width=\linewidth]{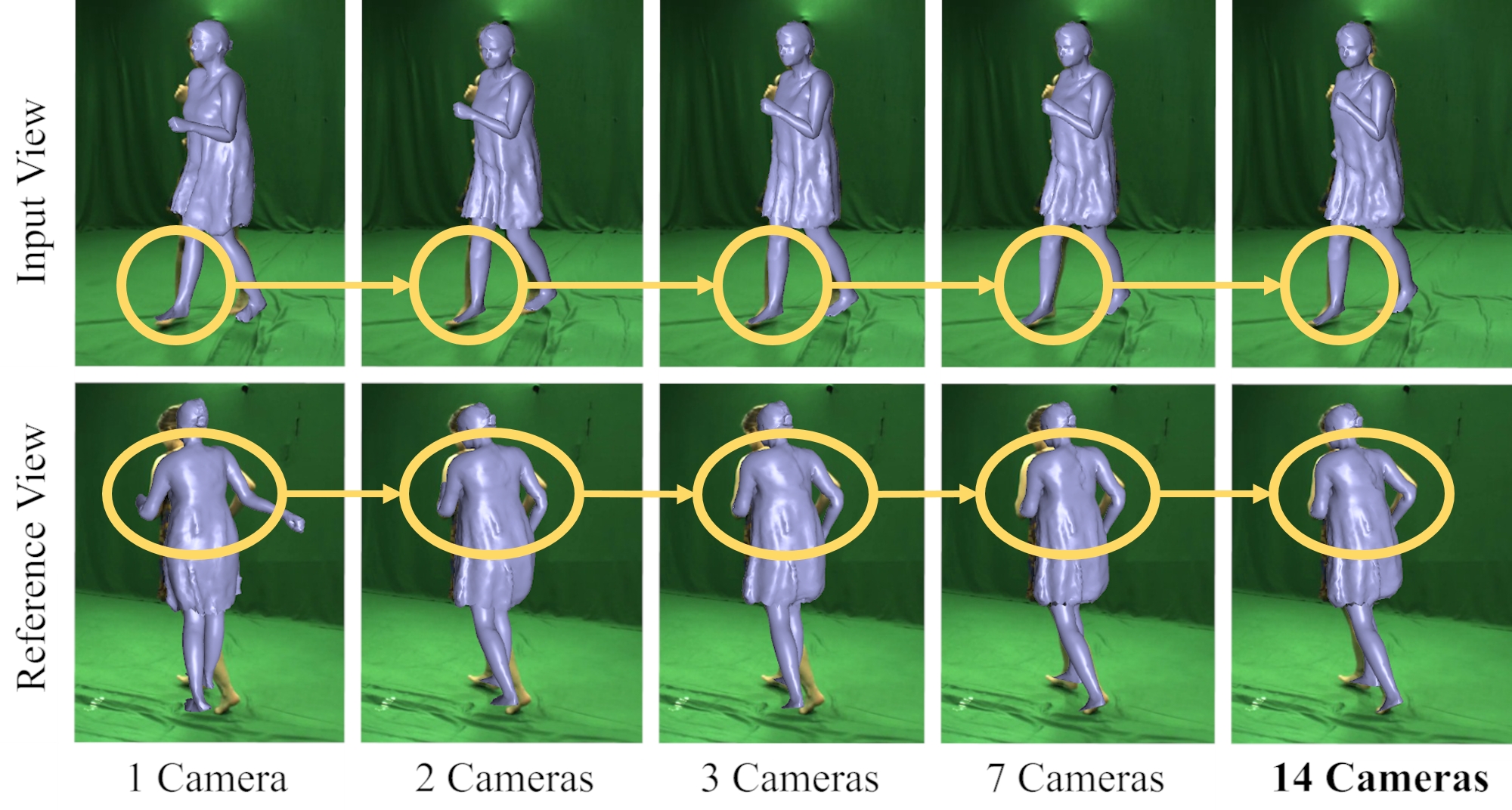}
	\end{center}
	\caption
	{
		Ablation for number of \emph{cameras} used during training.
		The most significant improvement happens when adding one additional camera to the monocular setting.
		But also adding further cameras consistently improves the result as the yellow circles indicate.
	}
	\label{fig:ablation_cameras}
\end{figure}

\begin{figure}[t]
	\begin{center}
		\includegraphics[width=\linewidth]{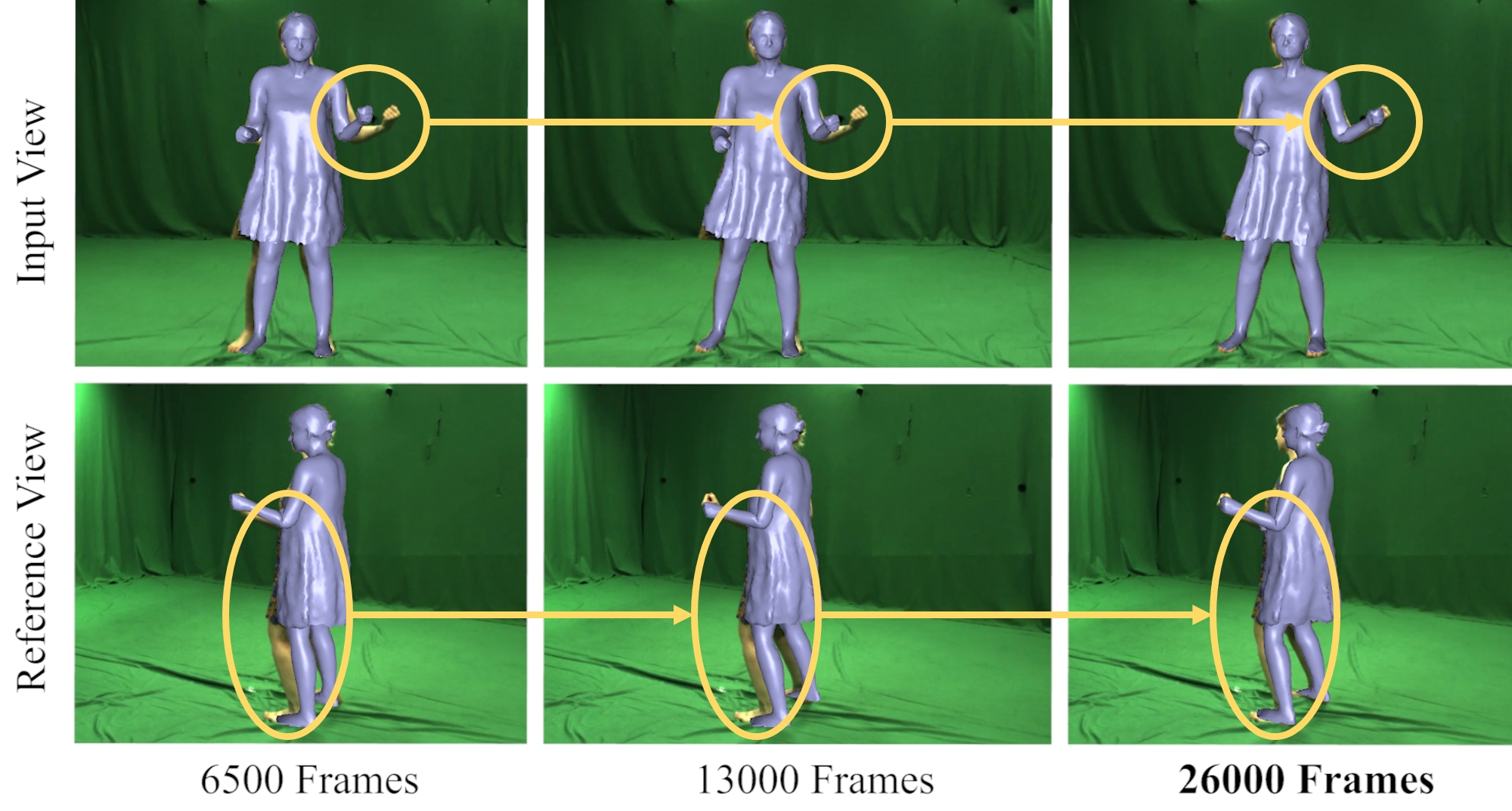}
	\end{center}
	\caption
	{
		Ablation for number of \emph{frames} used during training.
		The more frames we used during training the better the result becomes as the network can better sample the possible pose and deformation space.
	}
	\label{fig:ablation_frames}
\end{figure}

\begin{figure}[t]
	\begin{center}
		\includegraphics[width=\linewidth]{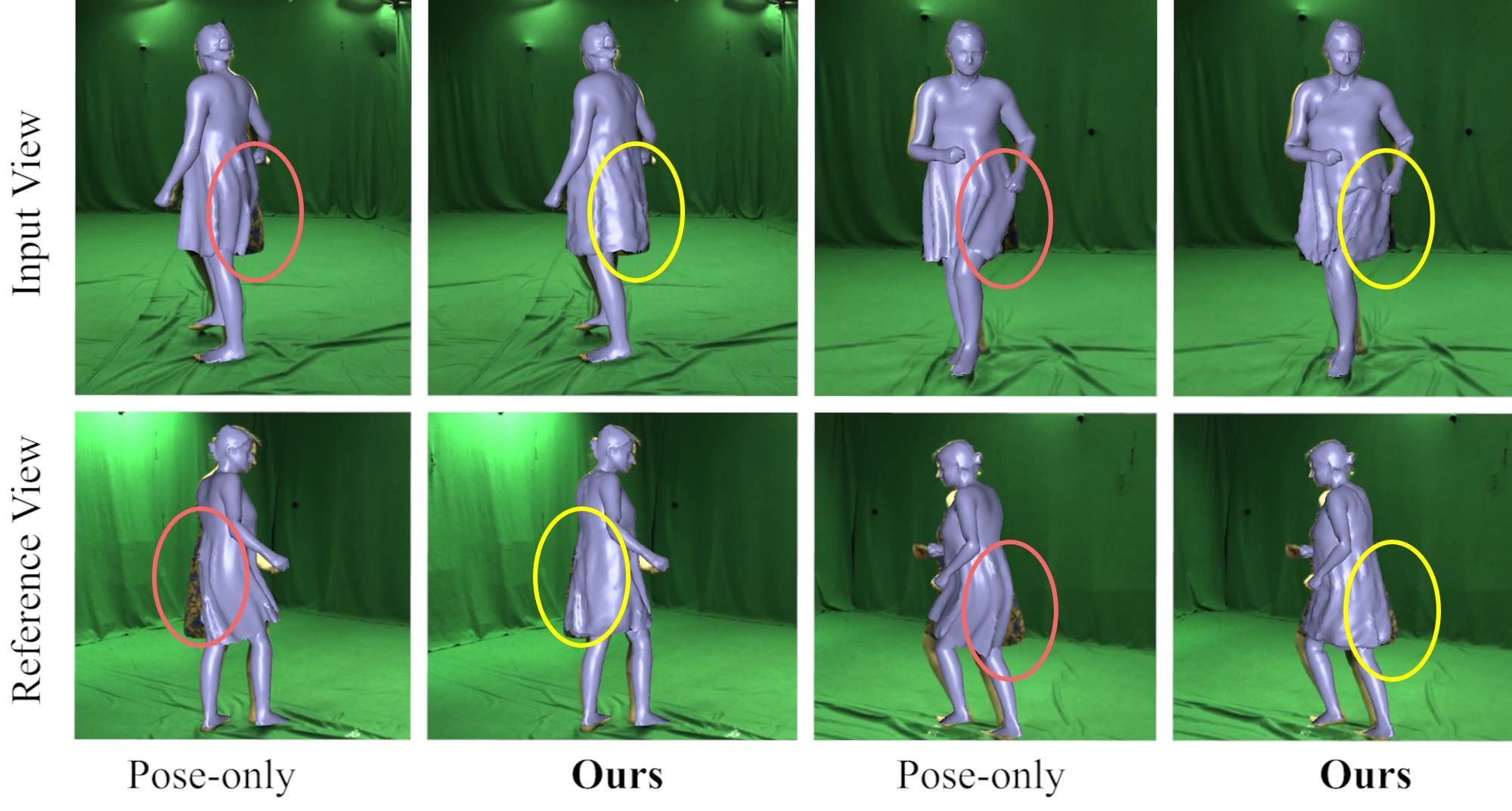}
	\end{center}
	\caption
	{
		Our result from the input view and a reference view that was not used for tracking.
		Note that our \textit{DefNet} can even regress deformations along the camera viewing axis of the input camera (second column) and it can correctly deform surface parts that are occluded (fourth column).
	}
	\label{fig:improvement_side}
\end{figure}

 \noindent \textbf{Qualitative Comparisons.}
Fig.~\ref{fig:qualitative} shows our qualitative results on in-the-wild test sequences with various clothing styles, poses and environments.
Our reconstructions not only precisely overlay with the input images, but also look plausible from arbitrary 3D view points.
In Fig.~\ref{fig:qualcomp} and \ref{fig:comparison1}, we qualitatively compare our approach to the related human capture and reconstruction methods~\cite{kanazawa18, habermann19, saito19, zheng19} on our green screen and the in-the-wild sequences, respectively.
In terms of the shape representation, our method is most closely related to LiveCap~\cite{habermann19} that also uses a person-specific template.
Since they non-rigidly fit the template only to the monocular input view, their results do not faithfully depict the deformation in other view points.
Further, their pose estimation severely suffers from the monocular ambiguities, whereas our pose results are more robust and accurate (see supplemental video).
Comparing to the other three methods~\cite{kanazawa18, saito19, zheng19} that are trained for general subjects, our approach has the following advantages:
First, our method recovers the non-rigid deformations of humans in general clothes whereas the parametric model-based approaches \cite{kanazawa18,kanazawa19} only recover naked body shape.
Second, our method directly provides surface correspondences over time which is important for AR/VR applications (see supplemental video).
In contrast, the results of implicit representation-based methods, PIFu~\cite{saito19} and DeepHuman~\cite{zheng19}, lack temporal surface correspondences and do not preserve the skeletal structure of the human body, \textit{i.e.}, they often exhibit missing arms and disconnected geometry.
Furthermore, DeepHuman~\cite{zheng19} only recovers a coarse shape in combination with a normal image of the input view, while our method can recover medium-level detailed geometry that looks plausible from all views.
Last but not least, all these existing methods have problems when overlaying their reconstructions on the reference view, even though some of the methods show a very good overlay on the input view.
In contrast, our approach reconstructs accurate 3D geometry, and therefore, our results can precisely overlay on the reference views (also see Fig.~\ref{fig:referenceview}, \ref{fig:ablation_cameras}, \ref{fig:ablation_frames}, and \ref{fig:improvement_side}).
\newcommand{\bmethodlabel}[2]{\scriptsize {\hspace{#1} {#2}}}	
\begin{figure}[t]
	\scriptsize
	\begin{center}
		\includegraphics[width=\linewidth]{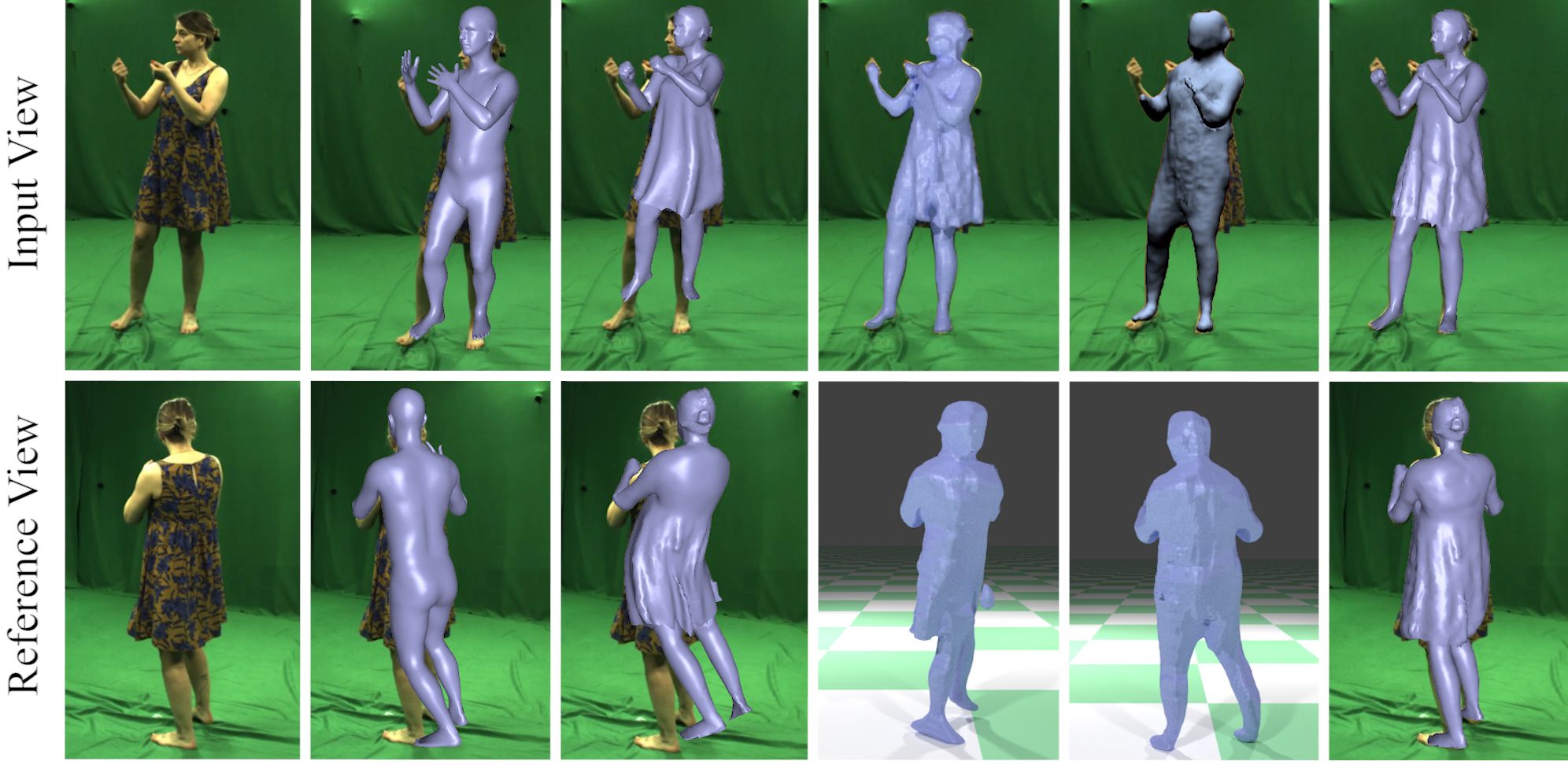}
	\end{center}
	\bmethodlabel{0.6cm}{Input}
	\bmethodlabel{0.4cm}{HMR~\cite{kanazawa18}}
	\bmethodlabel{0.1cm}{LiveCap~\cite{habermann19}}
	\bmethodlabel{0.1cm}{PIFu~\cite{saito19}}
	\bmethodlabel{0.01cm}{DeepHuman~\cite{zheng19}}
	\bmethodlabel{0.1cm}{\textbf{Ours}}
	\caption
	{
		Qualitative comparison to other methods \cite{kanazawa18,habermann19,saito19,zheng19} on our green screen evaluation sequences. 
		Note that our results overlay more accurately to the input view and also look more plausible from a reference view that was not used for tracking.
		Ground truth global translation is used to match the reference view for the results of~\cite{kanazawa18,habermann19}.
		Since PIFu~\cite{saito19} and DeepHuman~\cite{zheng19} output meshes with varying topology in a canonical volume without an attached root, it is not possible to apply the ground truth translation and therefore we show the reference view without overlay.
	}
	\label{fig:qualcomp}
\end{figure}
\begin{figure}[t]
	\scriptsize
	\begin{center}
		\includegraphics[width=\linewidth]{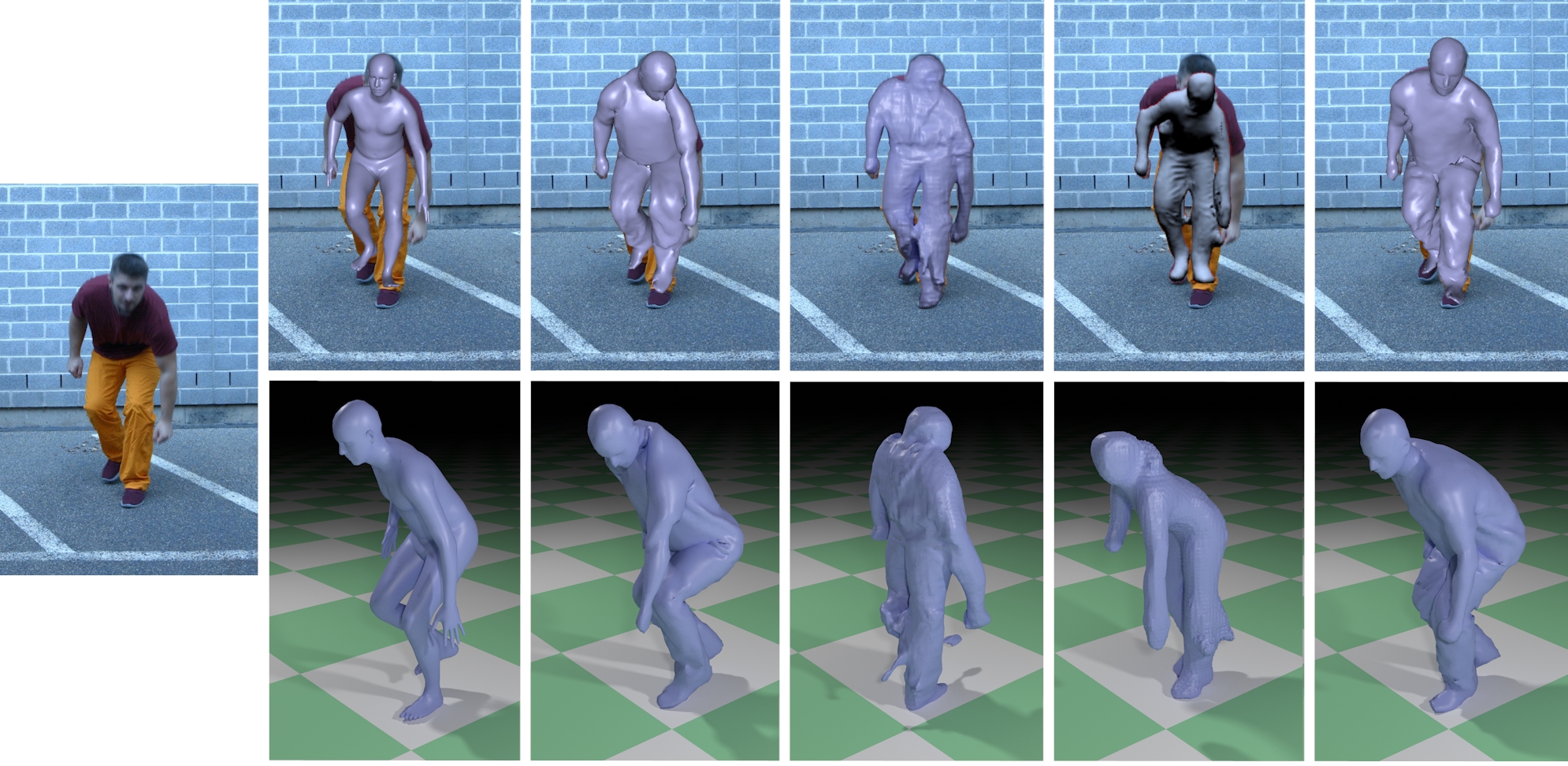}
	\end{center}
	\bmethodlabel{0.3cm}{Input}
	\bmethodlabel{0.6cm}{HMR~\cite{kanazawa18}}
	\bmethodlabel{0.25cm}{LiveCap~\cite{habermann19}}
	\bmethodlabel{0.22cm}{PIFu~\cite{saito19}}
	\bmethodlabel{0.01cm}{DeepHuman~\cite{zheng19}}
	\bmethodlabel{0.1cm}{\textbf{Ours}}
	\begin{center}
		\includegraphics[width=\linewidth]{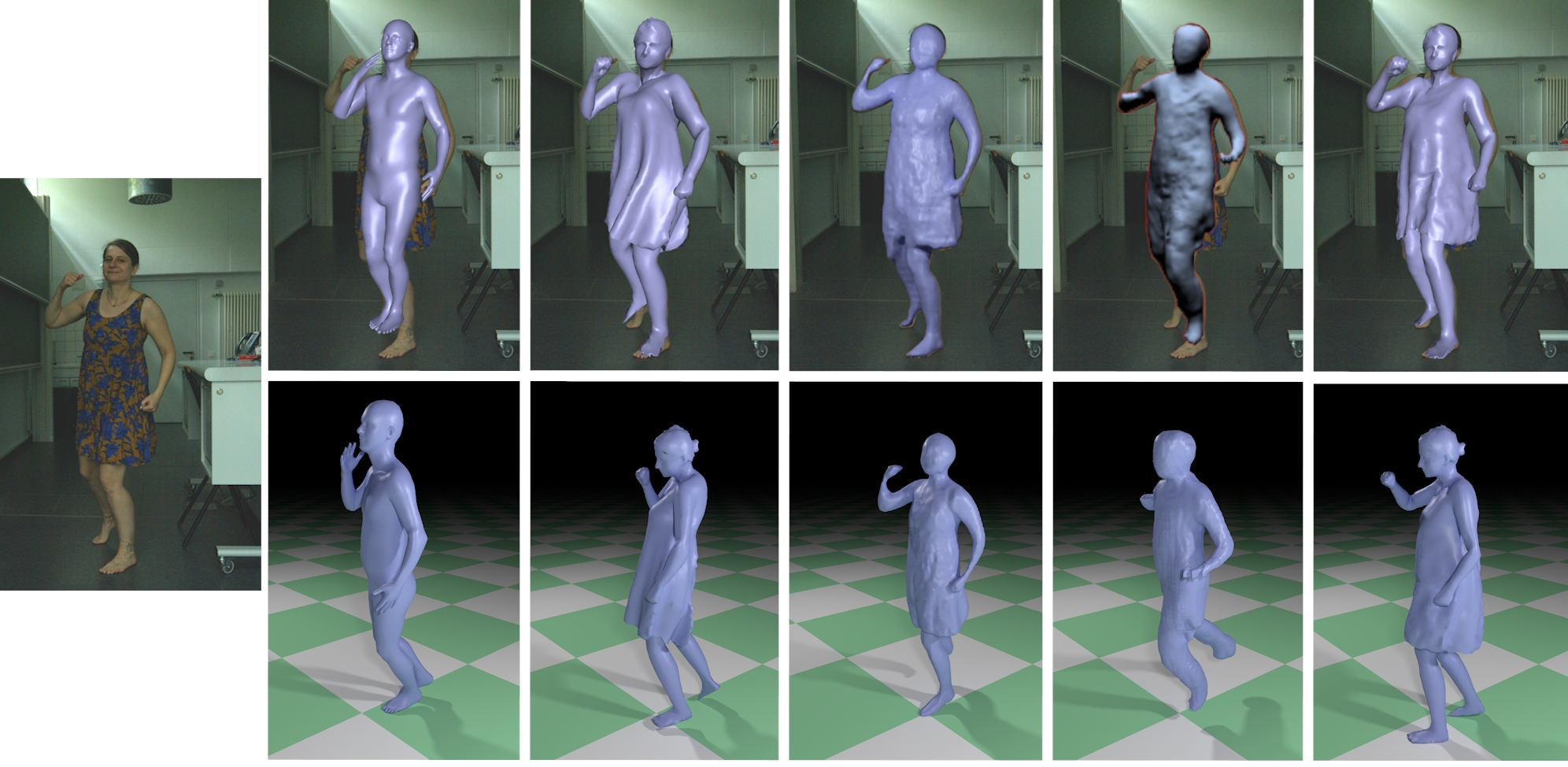}
	\end{center}
	\bmethodlabel{0.3cm}{Input}
	\bmethodlabel{0.6cm}{HMR~\cite{kanazawa18}}
	\bmethodlabel{0.25cm}{LiveCap~\cite{habermann19}}
	\bmethodlabel{0.22cm}{PIFu~\cite{saito19}}
	\bmethodlabel{0.01cm}{DeepHuman~\cite{zheng19}}
	\bmethodlabel{0.1cm}{\textbf{Ours}}
	
	\caption
	{
			Comparisons to related work \cite{kanazawa18, habermann19, saito19, zheng19} on our in-the-wild sequences showing \textit{S1} and \textit{S4}.
			Our approach can recover the deformations of clothing in contrast to \cite{kanazawa18} and gives more stable and accurate results in 3D compared to \cite{habermann19}. 
			Moreover, note that in contrast to previous work \cite{saito19, zheng19}, our method regresses space-time coherent geometry, which follows the structure of the human body.
	}
	\label{fig:comparison1}
\end{figure}
%
%
\par \noindent \textbf{Skeletal Pose Accuracy.}
We quantitatively compare our pose results (output of \textit{PoseNet}) to existing pose estimation methods on \textit{S1} to \textit{S4}.
To account for different types of apparel, we choose \textit{S1} and \textit{S2} wearing trousers and a T-shirt or a pullover and \textit{S3} and \textit{S4} wearing a long and short dress, respectively.
We rescale the bone length for all methods to the ground truth and evaluate the following metrics on the 14 commonly used joints~\cite{mehta17} for every 10th frame:
1) We evaluate the root joint position error or global localization error (\textit{GLE}) to measure how good the skeleton is placed in global 3D space.
Note that \textit{GLE} can only be evaluated for LiveCap~\cite{habermann19} and ours, since other methods only produce up-to-scale depth.
2) To evaluate the accuracy of the pose estimation, we report the 3D percentage of correct keypoints (3DPCK) with a threshold of $150mm$ of the root aligned poses and the area under the 3DPCK curve (AUC).
3) To factor out the errors in the global rotation, we also report the mean per joint position error (MPJPE) after Procrustes alignment.
We compare our approach against the state-of-the-art pose estimation approaches including VNect~\cite{mehta17}, HMR~\cite{kanazawa18}, HMMR~\cite{kanazawa19}, and LiveCap~\cite{habermann19}.
We also compare to a multi-view baseline approach (\textit{MVBL}), where we use our differentiable skeleton model in an optimization framework to solve for the pose per frame using the proposed multi-view losses.
We can see from Tab.~\ref{tab:ablationJoint} that our approach outperforms the related monocular methods in all metrics by a large margin and is even close to \textit{MVBL} although our method only takes a single image as input.
We further compare to VNect~\cite{mehta17} fine-tuned on our training images for \textit{S1}.
To this end, we compute the 3D joint position using The Captury~\cite{captury} to provide ground truth supervision for VNect.
On the evaluation sequence for \textit{S1}, the fine-tuned VNect achieved 95.66\% 3DPCK, 52.13\% AUC and 47.16$mm$ MPJPE.
This shows our weakly supervised approach yields comparable or better results than supervised methods in the person-specific setting.
However, our approach does not require 3D ground truth annotation that is difficult to obtain, even for only sparse keypoints, let alone the dense surfaces.
Further note that even for \textit{S3} we achieve accurate results even though she wears a long dress such that legs are mostly occluded.
On \textit{S2}, we found that our results are more accurate than \textit{MVBL} since the classical frame-to-frame optimization can get stuck in local minima, leading to wrong poses.
\begin{table}
	\begin{center}
		\begin{tabular}{|c|c|c|c|c|}
			\hline
			\multicolumn{5}{|c|}{\textit{MPJPE/GLE (in mm) and 3DPCK/AUC (in \%) on S1}} \\
			\hline
			\textbf{Method}                            & \textbf{GLE}$\downarrow$		& \textbf{3DPCK}$\uparrow$	& \textbf{AUC}$\uparrow$	 & \textbf{MPJPE}$\downarrow$		\\
			\hline
			VNect~\cite{mehta17} 			           & -   		  		& 66.06  	  		& 28.02			& 77.19		   		\\
			HMR~\cite{kanazawa18}		   	           & -      	  		& 82.39	  			& 43.61		    & 72.61				\\
			HMMR~\cite{kanazawa19}		   	           & -  	  	        & 87.48	  			& 45.33		    & 72.40				\\
			LiveCap~\cite{habermann19} 	               & 317.01		  		& 71.13	  			& 37.90	        & 92.84			 	\\
			Ours 									   & \textbf{91.08}		  		& \textbf{98.43}     		& \textbf{58.71}		    & \textbf{49.11}		      	\\
			\hline
			MVBL                                       & 76.03	      		& 99.17	  			& 57.79		    & 45.44				\\
			\hline
		\end{tabular}
	\end{center}
	\begin{center}
		\begin{tabular}{|c|c|c|c|c|}
			\hline
			\multicolumn{5}{|c|}{\textit{MPJPE/GLE (in mm) and 3DPCK/AUC (in \%) on S2}} \\
			\hline
			\textbf{Method}                            & \textbf{GLE}$\downarrow$		& \textbf{3DPCK}$\uparrow$	& \textbf{AUC}$\uparrow$	 & \textbf{MPJPE}$\downarrow$		\\
			\hline
			VNect~\cite{mehta17} 			           & -			  		& 80.50  	  		& 39.98			& 66.96		   		\\
			HMR~\cite{kanazawa18}		   	           & -		  	  		& 80.02	  			& 39.24		    & 71.87				\\
			HMMR~\cite{kanazawa19}		   	           & -       	  		& 82.08	  			& 41.00		    & 74.69				\\
			LiveCap~\cite{habermann19} 	               & 142.39		  		& 79.17	  			& 42.59	        & 69.18		 		\\
			Ours 									   & \textbf{75.79}		& \textbf{94.72}    & \textbf{54.61}& \textbf{52.71}	\\
			\hline
			MVBL                                       & 64.12	      		& 89.91	  			& 45.58		    & 57.52				\\
			\hline
		\end{tabular}
	\end{center}
	\begin{center}
		\begin{tabular}{|c|c|c|c|c|}
			\hline
			\multicolumn{5}{|c|}{\textit{MPJPE/GLE (in mm) and 3DPCK/AUC (in \%) on S3}} \\
			\hline
			\textbf{Method}                            & \textbf{GLE}$\downarrow$		& \textbf{3DPCK}$\uparrow$	& \textbf{AUC}$\uparrow$	 & \textbf{MPJPE}$\downarrow$		\\
			\hline
			VNect~\cite{mehta17} 			               				& -				  		& 78.03  	  			& 41.95			& 88.14		   		\\
			HMR~\cite{kanazawa18}		   	             				& -				  		& 83.37  	  			& 42.37			& 79.02		   		\\
			HMMR~\cite{kanazawa19}		   	             				& -		  	        	& 79.93  	  			& 36.27			& 91.62		     	\\
			LiveCap~\cite{habermann19} 	               					& 281.27		  		& 66.30  	  			& 31.44			& 98.76		    	\\
			Ours 									               	    & \textbf{89.54}		& \textbf{95.09}  	  	& \textbf{54.00}& \textbf{58.77}	\\
			\hline
			MVBL                                                    	& 67.82		  			& 96.37  	  			& 54.99			& 56.08	   			\\
			\hline
		\end{tabular}
	\end{center}
	\begin{center}
		\begin{tabular}{|c|c|c|c|c|}
			\hline
			\multicolumn{5}{|c|}{\textit{MPJPE/GLE (in mm) and 3DPCK/AUC (in \%) on S4}} \\
			\hline
			\textbf{Method}                            & \textbf{GLE}$\downarrow$		& \textbf{3DPCK}$\uparrow$	& \textbf{AUC}$\uparrow$	 & \textbf{MPJPE}$\downarrow$		\\
			\hline
			VNect~\cite{mehta17} 			               				& -     		  		& 82.06  	  			& 42.73			& 72.62		   		\\
			HMR~\cite{kanazawa18}		   	             				& -     		  		& 86.88  	  			& 43.91			& 73.63		   		\\
			HMMR~\cite{kanazawa19}		   	             				& -		  		        & 82.80  	  			& 41.18			& 77.41		   		\\
			LiveCap~\cite{habermann19} 	               					& 248.67		  		& 75.11  	  			& 37.35			& 83.48		   	    \\
			Ours 									               	    & \textbf{96.56}		  			& \textbf{96.74}  	  			& \textbf{59.25}			& \textbf{45.40}		   		\\
			\hline
			MVBL                                                    	& 75.82		  			& 96.20  	  			& 57.27			& 45.12		   		\\
			\hline
		\end{tabular}
	\end{center}
	\caption
	{
		Skeletal pose accuracy.
		Note that we are consistently better than other monocular approaches.
		Moreover, we are even close to the multi-view baseline.
	}
	\label{tab:joint}
\end{table}
%
%
\par \noindent \textbf{Surface Reconstruction Accuracy.}
To evaluate the accuracy of the regressed non-rigid deformations, we compute the intersection over union (IoU) between the ground truth foreground masks and the 2D projection of the estimated shape on \textit{S1} and \textit{S4} for every 100th frame.
We evaluate the IoU on \textit{all views}, on \textit{all views expect the input view}, and on the \textit{input view} which we refer to as \textit{AMVIoU}, \textit{RVIoU} and \textit{SVIoU}, respectively.
To factor out the errors in global localization, we apply the ground truth translation to the reconstructed geometries.
For DeepHuman~\cite{zheng19} and PIFu~\cite{saito19}, we cannot report the \textit{AMVIoU} and \textit{RVIoU}, since we cannot overlay their results on reference views as discussed before.
Further, PIFu~\cite{saito19} by design achieves perfect overlay on the input view, since they regress the depth for each foreground pixel.
However, their reconstruction does not reflect the true 3D geometry (see Fig.~\ref{fig:qualcomp}).
Therefore, it is meaningless to report their \textit{SVIoU}.
Similarly, DeepHuman~\cite{zheng19} achieves high \textit{SVIoU}, due to their volumetric representation.
But their results are often wrong, when looking from side views.
In contrast, our method consistently outperforms all other approaches in terms of \textit{AMVIoU} and \textit{RVIoU}, which shows the high accuracy of our method in recovering the 3D geometry.
Further, we are again close to the multi-view baseline.
\begin{table}
	\begin{center}
		\begin{tabular}{|c|c|c|c|}
			\hline
			\multicolumn{4}{|c|}{\textit{AMVIoU, RVIoU, and SVIoU (in \%) on S1 sequence}} \\
			\hline
			\textbf{Method}                                       	& \textbf{AMVIoU}$\uparrow$           	& \textbf{RVIoU}$\uparrow$     			& \textbf{SVIoU}$\uparrow$     		\\
			\hline
			HMR~\cite{kanazawa18}		                   			& 62.25		                          	& 61.7                       			& 68.85    							\\
			HMMR~\cite{kanazawa19}		                   			& 65.98 	                          	& 65.58                       			& 70.77   							\\
			LiveCap~\cite{habermann19} 	                 			& 56.02	                          		& 54.21                       			& 77.75	       	  					\\
			DeepHuman~\cite{zheng19} 		            			& -		                                & -                         			& \textbf{91.57}				      			\\
			Ours 									                & \textbf{87.2}	                      	 		& \textbf{87.03}                         		& 89.26      						\\
			\hline
			MVBL                                                    & 91.74		                       		& 91.72                         		& 92.02								\\
			\hline
		\end{tabular}
	\end{center}
	\begin{center}
		\begin{tabular}{|c|c|c|c|}
			\hline
			\multicolumn{4}{|c|}{\textit{AMVIoU, RVIoU and SVIoU (in \%) on S2}} \\
			\hline
			\textbf{Method}                                       	& \textbf{AMVIoU}$\uparrow$           	& \textbf{RVIoU}$\uparrow$     			& \textbf{SVIoU}$\uparrow$     			\\
			\hline
			HMR~\cite{kanazawa18}		                   			& 59.79		                            & 59.1                       			& 66.78    								\\
			HMMR~\cite{kanazawa19}		                   			& 62.64		                            & 62.03                       			& 68.77    								\\
			LiveCap~\cite{habermann19} 	                 			& 60.52	                          		& 58.82                       			& 77.75	       	  						\\
			DeepHuman~\cite{zheng19} 		            			& -		                                & -                         			& \textbf{91.57}				        \\
			Ours 									                & \textbf{83.73}	                   	& \textbf{83.49}                        & 89.26      							\\
			\hline
			MVBL                                                    & 89.62		                       		& 89.67                         		& 92.02									\\
			\hline
		\end{tabular}
	\end{center}
	\begin{center}
		\begin{tabular}{|c|c|c|c|}
			\hline
			\multicolumn{4}{|c|}{\textit{AMVIoU, RVIoU and SVIoU (in \%) on S3 }} \\
			\hline
			\textbf{Method}                                 		& \textbf{AMVIoU}$\uparrow$     		& \textbf{RVIoU}$\uparrow$     			& \textbf{SVIoU}$\uparrow$   			\\
			\hline
			HMR~\cite{kanazawa18}		              				& 59.05		               				& 58.73                         		& 63.12		                			\\
			HMMR~\cite{kanazawa19}		              				& 61.73		               				& 61.32                         		& 67.14		                			\\
			LiveCap~\cite{habermann19} 	            				& 61.55                              	& 60.47                         		& 75.6		                   			\\
			DeepHuman~\cite{zheng19} 		        				& -		                            	& -                                 	& 79.66				                	\\
			Ours 									               	& \textbf{85.75}	                    & \textbf{85.55}                        & \textbf{88.27}			            \\
			\hline
			MVBL                                                  	& 90.31		                    		& 90.21                          		& 91.53				 	            	\\
			\hline
		\end{tabular}
	\end{center}
	\begin{center}
		\begin{tabular}{|c|c|c|c|}
			\hline
			\multicolumn{4}{|c|}{\textit{AMVIoU, RVIoU, and SVIoU (in \%) on S4 sequence}} \\
			\hline
			\textbf{Method}                                 		& \textbf{AMVIoU}$\uparrow$            & \textbf{RVIoU}$\uparrow$     			& \textbf{SVIoU}$\uparrow$     	   	\\
			\hline
			HMR~\cite{kanazawa18}		              				& 65.1		                           & 64.66                         			& 70.84		                		\\
			HMMR~\cite{kanazawa19}		              				& 63.79		                           & 63.29                         			& 70.23		                		\\
			LiveCap~\cite{habermann19} 	            				& 59.96	                               & 59.02                         			& 72.16		                   		\\
			DeepHuman~\cite{zheng19} 		        				& -		                               & -                                 		& 84.15				                \\
			Ours 									               	& \textbf{82.53}		                           & \textbf{82.22}                          		& \textbf{86.66}			               		\\
			\hline
			MVBL                                                  	& 88.14		                           & 88.03                          		& 89.66				 	            \\
			\hline
		\end{tabular}
	\end{center}
	\caption
	{
		Surface deformation accuracy.
		Note that we again outperform all other monocular methods and are close to the multi-view baseline.
		Further note, that for \cite{zheng19} an evaluation of the multi-view IoU is not possible since their output is always in local image space that cannot be brought to global space.
	}
	\label{tab:surface}
\end{table}
%
%
\par
\begin{figure}[t]
	\begin{center}
		\includegraphics[width=\linewidth]{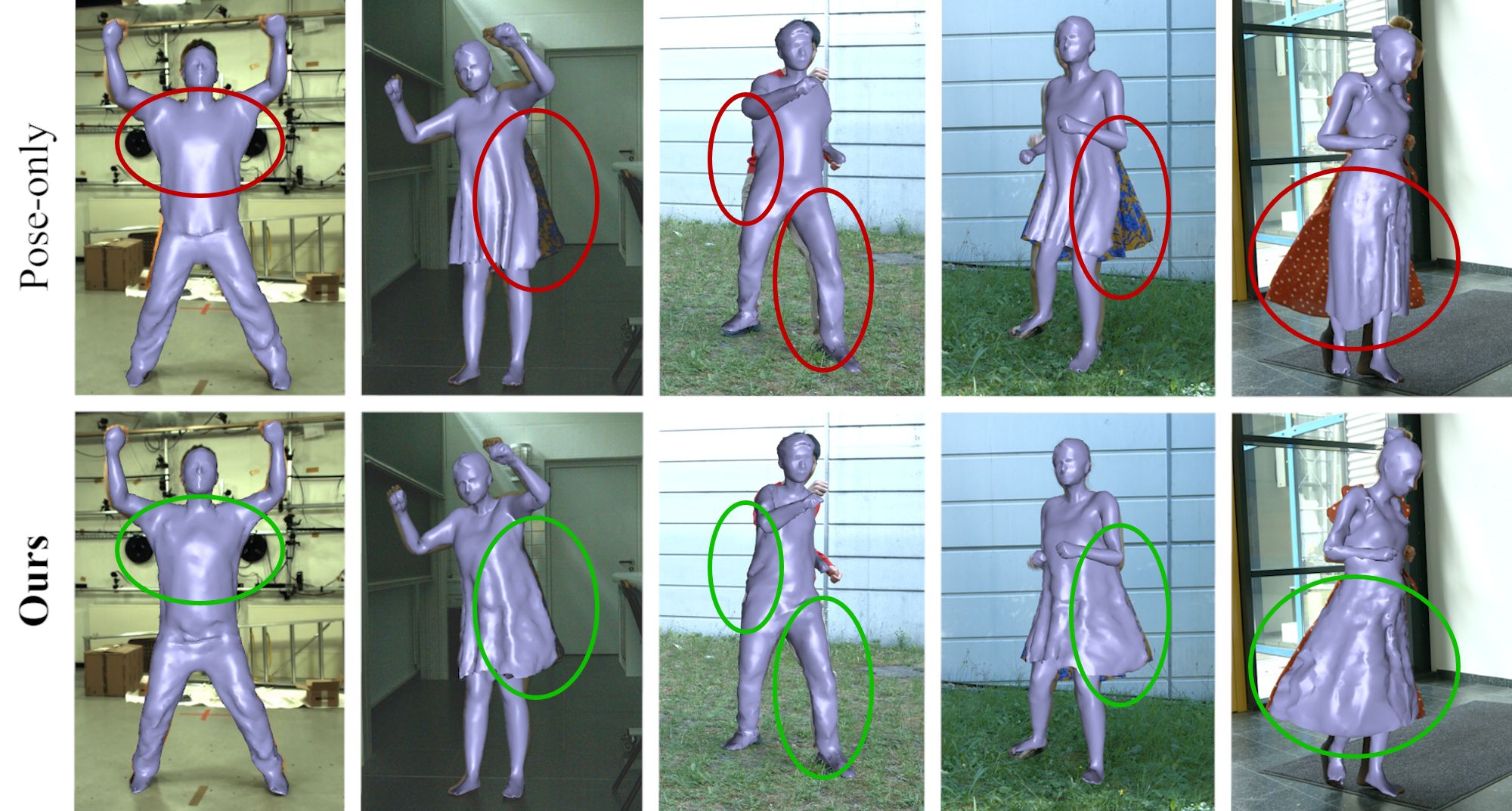}
	\end{center}
	\caption
	{
		\textit{PoseNet} + \textit{DefNet} vs. \textit{PoseNet-only}.
		\textit{DefNet} can deform the template to accurately match the input, especially for loose clothing.
		In addition, \textit{DefNet} also corrects slight errors in the pose and typical skinning artifacts.
	}
	\label{fig:nrvspose}
\end{figure}

\begin{figure}[t]
	\begin{center}
		\includegraphics[width=\linewidth]{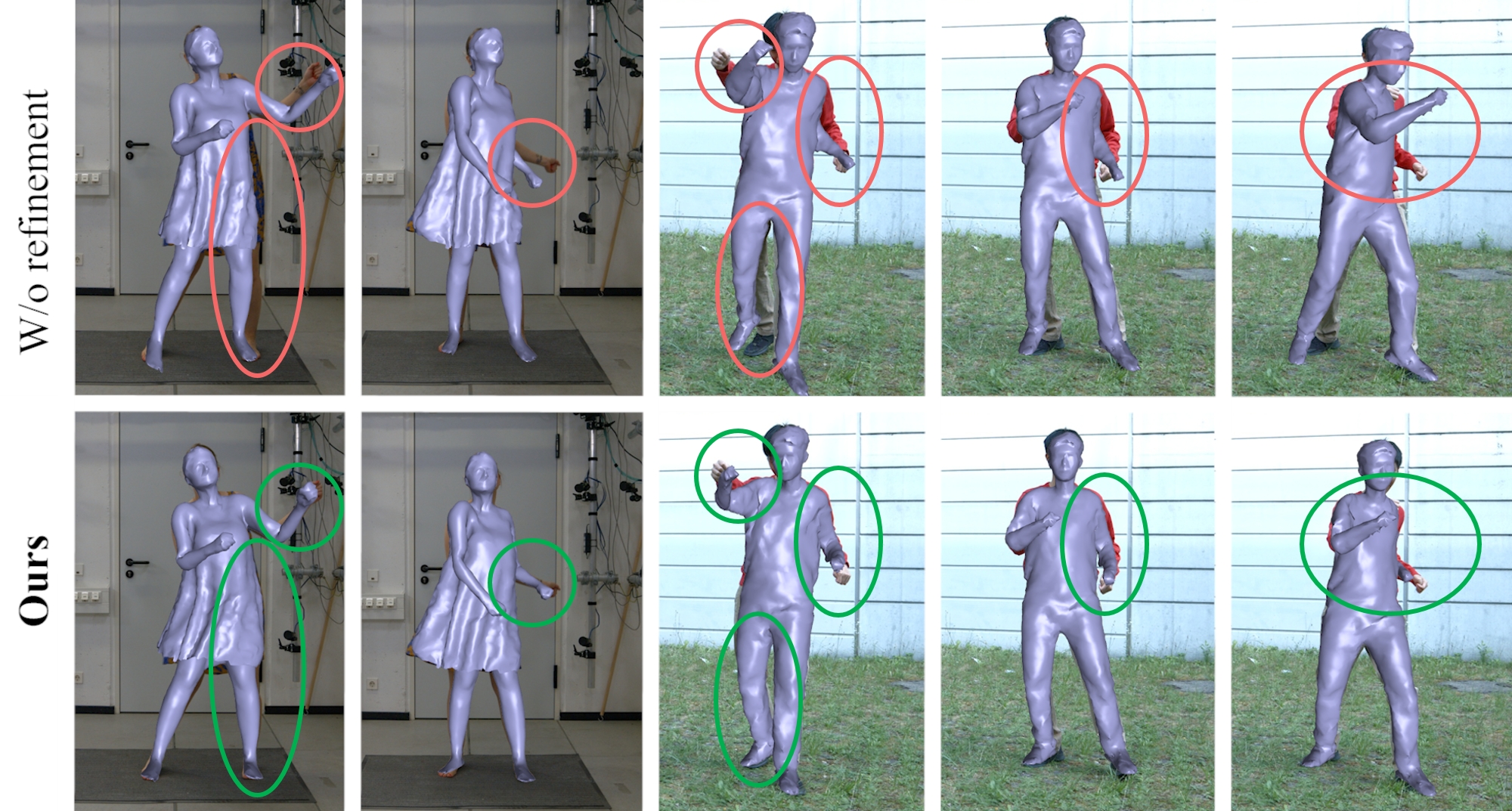}
	\end{center}
	\caption
	{
		Impact of the in-the-wild domain adaption step.
		Note that after the network refinement, both, the pose as well as the deformations better match the input.
	}
	\label{fig:refinement}
\end{figure}

 \noindent \textbf{Ablation Study.}
To evaluate the importance of the number of cameras, the number of training images, and our \textit{DefNet}, we performed an ablation study on \textit{S4} in Tab.~\ref{tab:ablationJoint}. 
1) In the first group of Tab.~\ref{tab:ablationJoint}, we train our networks with supervision using 1 to 14 views.
We can see that adding more views consistently improves the quality of the estimated poses and deformations.
The most significant improvement is from one to two cameras.
This is not surprising, since the single camera settings is inherently ambiguous.
In Fig.~\ref{fig:ablation_cameras}, the importance of the number of cameras is also shown qualitatively.
2) In the second group of Tab.~\ref{tab:ablationJoint} and in Fig.~\ref{fig:ablation_frames}, we reduce the training data to 1/2 and 1/4.
We can see that the more frames with different poses and deformations are seen during training, the better the reconstruction quality is.
This is expected since a larger number of frames may better sample the possible space of poses and deformations.
3) In the third group of Tab.~\ref{tab:ablationJoint}, we evaluate the \textit{AMVIoU} on the template mesh animated with the results of \textit{PoseNet}, which we refer to as \textit{PoseNet-only}.
One can see that on average, the \textit{AMVIoU} is improved by around 4\%.
Since most non-rigid deformations rather happen locally, the difference is visually even more significant as shown in Fig.~\ref{fig:nrvspose}.
Especially, the skirt is correctly deformed according to the input image whereas the \textit{PoseNet-only} result cannot fit the input due to the limitation of skinning.
In Fig.~\ref{fig:improvement_side}, we show the \textit{PoseNet-only} result and our final result on one of our evaluation sequences where a reference view is available.
The deformed template also looks plausible from a reference view that was not used for tracking.
Importantly, \textit{DefNet} can correctly regress deformations that are along the camera viewing direction of the input camera (see reference view in second column) and surface parts that are even occluded (see reference view in fourth column).
This implies that our weak multi-view supervision during training let the network learn the entire 3D surface deformation of the human body.
For more results, we refer to the supplemental video.
\begin{table}
	\begin{center}
		\begin{tabular}{|c|c|c|}
			\hline
			\multicolumn{3}{|c|}{\textit{3DPCK and AMVIoU (in \%) on S4 sequence}} \\
			\hline
			\textbf{Method}                                     & \textbf{3DPCK}$\uparrow$   	& \textbf{AMVIoU}$\uparrow$    	\\
			\hline
			1 camera view 									        & 62.11		       			& 65.11		    				\\
			2 camera views 									        & 93.52	           			& 78.44	   						\\
			3 camera views 									        & 94.70		           		& 79.75		   					\\
			7 camera views 										    & 95.95		           		& 81.73							\\
			\hline
			6500 frames 									   	& 85.19		        	 	& 73.41		  					\\
			13000 frames 									   	& 92.25		        		& 78.97		     				\\
			\hline
			PoseNet-only                                        & 96.74	            		& 78.51		   					\\
			Ours(14 views, 26000 frames) 				  			& \textbf{96.74}		   	& \textbf{82.53}				\\
			\hline
		\end{tabular}
	\end{center}
	\caption
	{
		Ablation study.
		We evaluate the number of cameras and the number of frames used during training in terms of the \textit{3DPCK} and \textit{AMVIoU} metrics.	
		Adding more cameras and frames consistently improves the quality of reconstruction.
		Further, \textit{DefNet} improves the \textit{AMVIoU} compared to pure pose estimation.
	}
	\label{tab:ablationJoint}
\end{table}
4) finally, in Fig.~\ref{fig:refinement}, we visually demonstrate the impact of our domain adaptation step.
It becomes obvious that the refinement drastically improves the pose as well as the non-rigid deformations so that the input can be matched at much higher accuracy.
Further, we do not require any additional input for the refinement as our losses can be directly adapted to the monocular setting.
%
%
\par
\begin{figure}[t]
	\begin{center}
		\includegraphics[width=\linewidth]{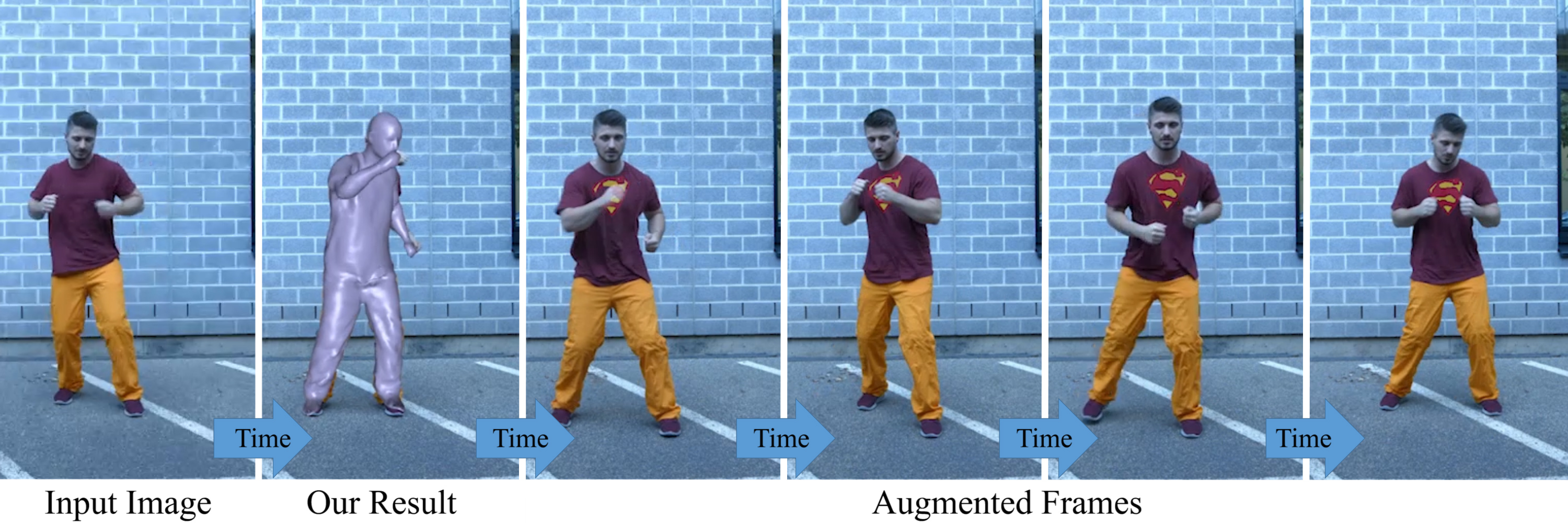}
	\end{center}
	\caption
	{
		Video augmentation.
		Our method can be used to augment a video with textures like the logo on the T-shirt.
		Since we track the underlying 3D geometry our method also accounts for occlusions of the augmented texture.
	}
	\label{fig:applications}
\end{figure}

 \noindent \textbf{Applications.}
Our method enables driving 3D characters just from a monocular RGB video (see Fig.~\ref{fig:qualitative}).
As the only device that is needed at test time is a single color camera, our method can be easily used in daily life scenarios once the multi-view video and the template are acquired and the training of the model was performed.
Further, as we also account for non-rigid surface deformations, our method also enhances the realism of the virtual characters.
Our approach also allows augmenting a video as shown in Fig.~\ref{fig:applications}.
Since we track the entire 3D geometry, the augmented texture is also aware of occlusions in contrast to pure image-based augmentation techniques.
\section{Conclusion}
\label{sec:conclusion}
\par \noindent \textbf{Limitations.}
Conceptually, both representations, pose and the non-rigid deformations, are decoupled.
Nevertheless, since the predicted poses during training are not perfect, our \textit{DefNet} also deforms the graph to account for wrong poses to a certain degree.
In our supplemental video, we also tested our method on subjects that were not used for training but which wear the same clothing as the training subject.
Although, our method still performs reasonable, the overall accuracy drops as the subjects appearance was never observed during training.
Further, our method can fail for extreme poses, e.g. a hand stand, that were not observed during training.
%
%
\par
We have presented a learning-based approach for monocular dense human performance capture using only weak multi-view supervision. 
In contrast to existing methods, our approach directly regresses poses and surface deformations from neural networks, produces temporal surface correspondences, preserves the skeletal structure of the human body, and can handle loose clothes.
Our qualitative and quantitative results in different scenarios show that our method produces more accurate 3D reconstruction of pose and non-rigid deformation than existing methods.
In the future, we plan to incorporate hands and the face to our mesh representation to enable joint tracking of body, facial expressions and hand gestures.
We are also interested in physically more correct multi-layered representations to model the garments even more realistically.

 \section*{Acknowledgment}
This work was funded by the ERC Consolidator Grant 4DRepLy (770784) and the Deutsche Forschungsgemeinschaft (Project Nr. 409792180, Emmy Noether Programme, project: Real Virtual Humans).

\ifCLASSOPTIONcaptionsoff
  \newpage
\fi

{\small
	\bibliographystyle{IEEEtran}
	\bibliography{bibliography}
}

%

\begin{IEEEbiography}[{\includegraphics[width=1in,height=1.25in,clip,keepaspectratio]{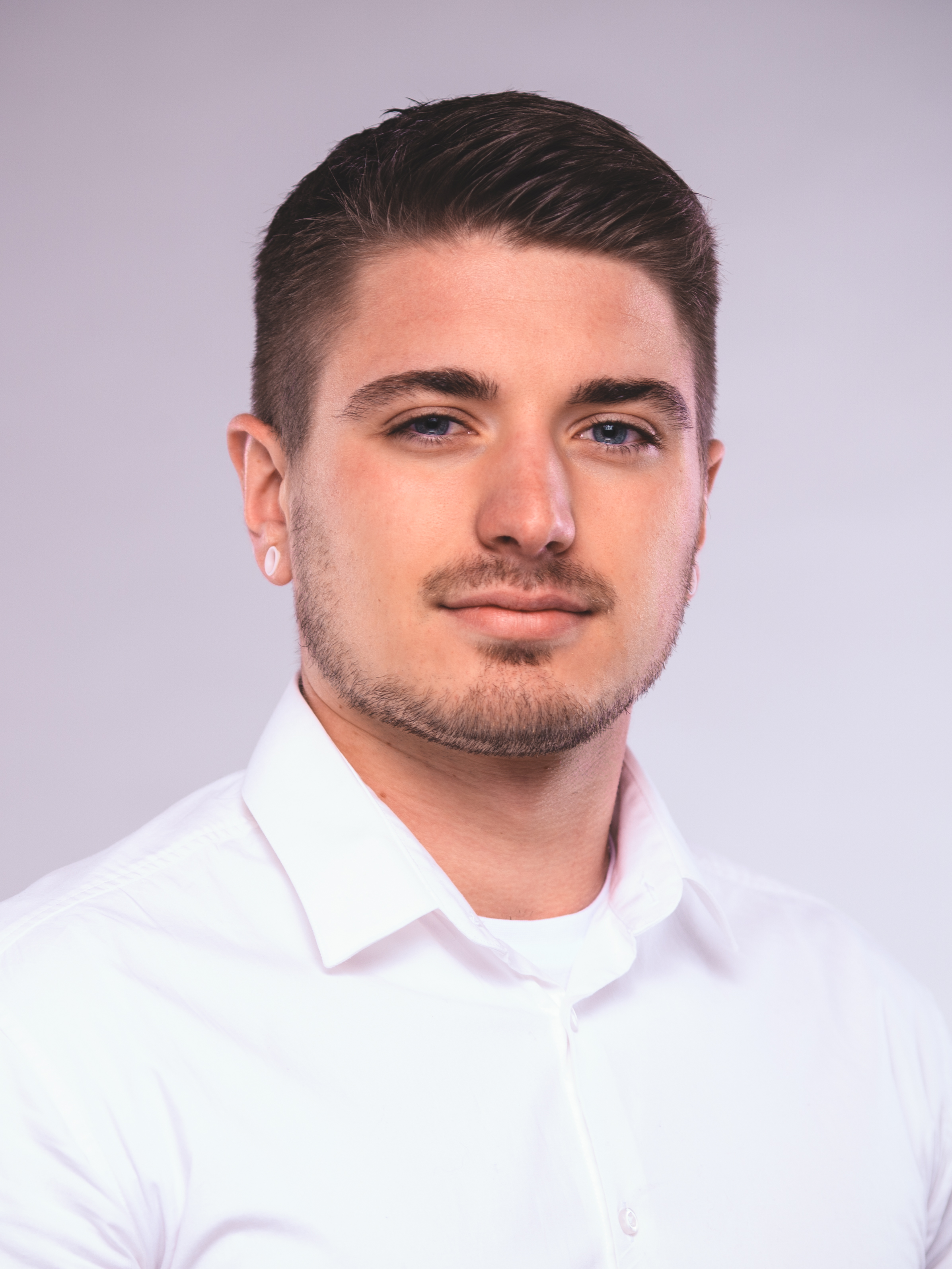}}]{Marc Habermann} works as a PhD student in the Graphics, Vision and Video group at the Max Planck Institute for Informatics. Within his thesis, he explores the modeling and tracking of non-rigid deformations of surfaces, e.g. capturing the performance of humans in their everyday clothing. In his previous works, he showed that this is possible at real-time frame rates and that the 3D performance can be further improved using deep learning techniques. He received the Guenter-Hotz-Medal for the best Master graduates in Computer Science at Saarland University in 2017 and his work, DeepCap, received the CVPR Best Student Paper Honorable Mention in 2020.
\end{IEEEbiography}

\begin{IEEEbiography}[{\includegraphics[width=1in,height=1.25in,clip,keepaspectratio]{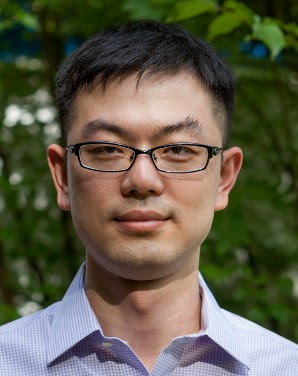}}]{Weipeng Xu} is a research scientist at Facebook Reality Labs in Pittsburgh. He was a post-doctoral researcher at the Graphic, Vision \& Video group of Max Planck Institute for Informatics in Saarbruecken, Germany. He received B.E. and Ph.D. degrees from Beijing Institute of Technology in 2009 and 2016, respectively. He studied as a long-term visiting student at NICTA and Australian National University from 2013 to 2015. His research interests include virtual human character, human pose estimation and machine learning for vision/graphics.
\end{IEEEbiography}

\begin{IEEEbiography}[{\includegraphics[width=1in,height=1.25in,clip,keepaspectratio]{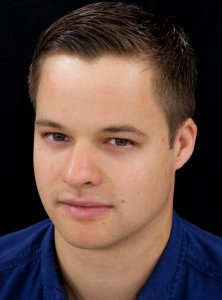}}]{Michael Zollhoefer} is a Visiting Assistant Professor at Stanford University. His stay at Stanford is funded by a postdoctoral fellowship of the Max Planck Center for Visual Computing and Communication (MPC-VCC), which he received for his work in the fields of computer vision, computer graphics, and machine learning. Before, Michael was a Postdoctoral Researcher in the Graphics, Vision \& Video group at the Max Planck Institute for Informatics in Saarbruecken, Germany. He received his PhD in 2014 from the University of Erlangen-Nuremberg for his work on real-time static and dynamic scene reconstruction. His research is focused on teaching computers to reconstruct and analyze our world at frame rate based on visual input. To this end, he develops key technology to invert the image formation models of computer graphics based on data-parallel optimization and state-of-the-art deep learning techniques.
\end{IEEEbiography}

\begin{IEEEbiography}[{\includegraphics[width=1in,height=1.25in,clip,keepaspectratio]{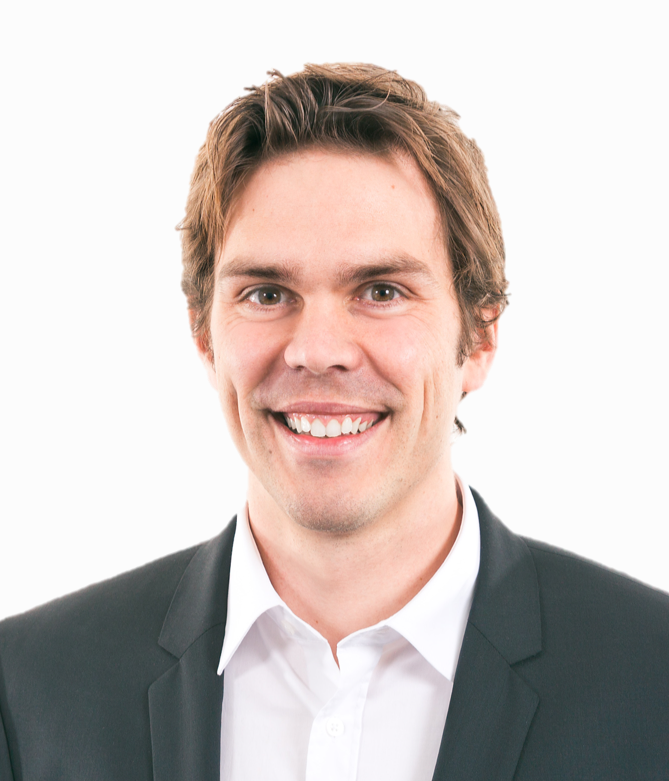}}]{Gerard Pons-Moll} is the head of the Emmy Noether independent research group "Real Virtual Humans", senior researcher at the Max Planck for Informatics (MPII) in Saarbrücken, Germany, and Junior Faculty at Saarland Informatics Campus. His research lies at the intersection of computer vision, computer graphics and machine learning -- with special focus on analyzing people in videos, and creating virtual human models by "looking" at real ones. His research has produced some of the most advanced statistical human body models of pose, shape, soft-tissue and clothing (which are currently used for a number of applications in industry and research), as well as algorithms to track and reconstruct 3D people models from images, video, depth, and IMUs. His work has received several awards including the prestigious Emmy Noether Grant (2018), a Google Faculty Research Award (2019), a Facebook Reality Labs Faculty Award (2018), and recently the German Pattern Recognition Award (2019), which is given annually by the German Pattern Recognition Society to one outstanding researcher in the fields of Computer Vision and Machine Learning. In 2020 he received a Snap-Research gift. His work got Best Papers Awards BMVC’13, Eurographics’17, 3DV'18 and CVPR'20 and has been published at the top venues and journals including CVPR, ICCV, Siggraph, Eurographics, 3DV, IJCV and PAMI. He served as Area Chair for ECCV'18, 3DV'19, SCA'18'19, FG'20, ECCV'20. He will serve as Area Chair for CVPR'21 and 3DV'20.
\end{IEEEbiography}

\begin{IEEEbiography}[{\includegraphics[width=1in,height=1.25in,clip,keepaspectratio]{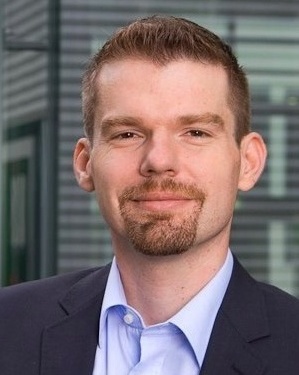}}]{Christian Theobalt} is a Professor of Computer Science and the head of the research group “Graphics, Vision, \& Video” at the Max-Planck-Institute for Informatics, Saarbruecken, Germany. He is also a professor at Saarland University. His research lies on the boundary between Computer Vision and Computer Graphics. For instance, he works on 4D scene reconstruction, marker-less motion and performance capture, machine learning for graphics and vision, and new sensors for 3D acquisition. Christian received several awards, for instance the Otto Hahn Medal of the Max-Planck Society (2007), the EUROGRAPHICS Young Researcher Award (2009), the German Pattern Recognition Award (2012), an ERC Starting Grant (2013), an ERC Consolidator Grant (2017), and the Eurographics Outstanding Technical Contributions Award (2020). In 2015, he was elected one of Germany’s top 40 innovators under 40 by the magazine Capital. He is a co-founder of theCaptury (www.thecaptury.com).
\end{IEEEbiography}

\vfill


\end{document}